\documentclass[10pt]{article} 

\usepackage[accepted]{rlj} 
\usepackage{nicefrac}

\usepackage{amssymb}            
\usepackage{mathtools}          
\usepackage{mathrsfs}           
\usepackage{graphicx}           
\usepackage{wrapfig}  
\usepackage{booktabs}
\usepackage{float}
\usepackage{threeparttable}
\usepackage{multirow} 
\usepackage{amsmath}        
\usepackage[mathscr]{eucal}
\usepackage{xcolor}
\usepackage{subcaption}         
\usepackage[space]{grffile}     
\usepackage{url}                
\usepackage{lipsum}             

\title{The Cell Must Go On: Agar.io for \\Continual Reinforcement Learning}

\setrunningtitle{The Cell Must Go On: Agar.io for Continual Reinforcement Learning}

\author{Mohamed A. Mohamed$^{1,2}$, Kateryna Nekhomiazh$^{3}$, Vedant Vyas$^{1}$, \\Marcos M. Jos\'e$^{1, 2}$, Andrew Patterson$^{4}$, Marlos C. Machado$^{1,2,5}$ }

\coverPageAuthor{Mohamed A. Mohamed, Kateryna Nekhomiazh, Vedant Vyas, \\Marcos M. Jos\'e, Andrew Patterson, Marlos C. Machado }

\emails{}

\affiliations{
$^{1}$\textbf{University of Alberta} \hspace{1cm}
$^{2}$\textbf{Alberta Machine Intelligence Institute (Amii)}\\ $^{3}$\textbf{NVIDIA} \hspace{2.9cm} $^{4}$\textbf{RLCore} \hspace{1.4cm} $^{5}$\textbf{Canada CIFAR AI Chair}\\
\par 
}
\contribution{
    We introduce \texttt{AgarCL}, a non-episodic, high-dimensional reinforcement learning (RL) environment based on the game Agar.io with non-stationary dynamics and a hybrid action space.
}
{
    \textsc{GoBigger}~\citep{zhang2023gobigger} is the only existing Agar.io-based platform, but it was designed for \emph{episodic} multi-agent RL. Earlier work explored Agar-style games in much simpler settings~\citep{anso2019deep,wiehe2018sampled}. \texttt{AgarCL} is much faster than alternative platforms, has different observation streams, and is well-suited to continual RL. \looseness=-1
}

\contribution{
    We introduce a suite of diagnostic mini-games in \texttt{AgarCL} that isolate specific challenges.
}
{
    Mini-games allow us to focus on specific challenges posed by \texttt{AgarCL} and to benchmark agents with respect to the particular skill a mini-game requires. They enable faster scientific iteration and may be viewed as intermediate steps toward an effective agent.
    
}

\contribution{
    In one setting where agents are able to learn to maximize their return, we show that the performance of a deployed fixed policy degrades over time in \texttt{AgarCL}.
}
{
    The policies learned by PPO, our best-performing agent, that were checkpointed at 32M and 48M steps, collapse over time when deployed in \texttt{AgarCL}. We see such a collapse in more challenging settings with more non-stationarity, not in the easier settings. 
}

\contribution{
    We evaluate DQN, PPO, and SAC on the full-game setting and across all the mini-games. We also evaluated PPO with GRUs, Continual Backprop, ReDo, and Shrink \& Perturb, and we observed that they were also unable to achieve strong performance in \texttt{AgarCL}. 
}
{
    We considered multiple settings, and the different combinations were evaluated in specific settings. We generally report the performance over 10 independent runs after hyperparameter tuning (when appropriate). These results are intended to characterize the difficulty of the environment rather than to establish state-of-the-art algorithmic performance. Results for the continual learning baselines were obtained with standard hyperparameters.
}


\summary{
Continual reinforcement learning (RL) concerns agents that are expected to learn continually, rather than converge to a policy that is then fixed for evaluation. This setting is well-suited to environments that the agent perceives as \textit{changing} over time, rendering any static policy ineffective. In continual RL, researchers often simulate such changes either by modifying episodic environments to incorporate task shifts during interaction or by designing simulators that explicitly model continual dynamics. However, transforming episodic problems into continual ones primarily captures scenarios involving abrupt changes in the data stream and still relies on episodic structure. Meanwhile, the few simulators explicitly designed for empirical continual RL research are often limited in scope or complexity. In this paper, we introduce \texttt{AgarCL}, a research platform for continual RL that enables agents to progress toward increasingly sophisticated behaviour. \texttt{AgarCL} is based on the game Agar.io, a non-episodic, high-dimensional problem with stochastic, ever-evolving dynamics, continuous actions, and partial observability. The contribution(s) box below describes our main empirical contributions in more detail.
}

\begin{document}

\makeCover  

\maketitle  

\begin{abstract}
Continual reinforcement learning (RL) concerns agents that are expected to learn continually, rather than converge to a policy that is then fixed for evaluation. This setting is well-suited to environments that the agent perceives as \textit{changing} over time, rendering any static policy ineffective. In continual RL, researchers often simulate such changes either by modifying episodic environments to incorporate task shifts during interaction or by designing simulators that explicitly model continual dynamics. However, transforming episodic problems into continual ones primarily captures scenarios involving abrupt changes in the data stream and still relies on episodic structure. Meanwhile, the few simulators explicitly designed for empirical continual RL research are often limited in scope or complexity. In this paper, we introduce \texttt{AgarCL}, a research platform for continual RL that enables agents to progress toward increasingly sophisticated behaviour. \texttt{AgarCL} is based on the game Agar.io, a non-episodic, high-dimensional problem with stochastic, ever-evolving dynamics, continuous actions, and partial observability. We provide benchmark results for DQN, PPO, and SAC on the primary continual RL challenge, as well as across a suite of smaller tasks within \texttt{AgarCL}. These smaller tasks isolate aspects of the full environment and allow us to characterize the distinct challenges posed by different components of the game. We further evaluate three continual learning methods—Shrink and Perturb, ReDo, and Continual Backpropagation—and observe they are also unable to achieve strong performance, suggesting that the challenges posed by \texttt{AgarCL} extend beyond the stability–plasticity dilemma. \looseness=-1
\end{abstract}


\section{Introduction}
\label{sec_introduction}

Continual reinforcement learning (RL) is the setting in which the agent is expected to learn continually, rather than converge to a policy that is subsequently fixed for evaluation or deployment. It can be viewed either as a problem formulation~\citep{khetarpal2022continualreinforcementlearningreview,abel2023definition,kumar2023continual} or as a solution method for problems perceived by the agent as non-stationary. Such scenarios are often motivated by the big world hypothesis, which posits that the world is ``bigger'' than the agent~\citep{javed2024big,lewandowski2025world}. In this regime, continual adaptation is more effective than committing to a static policy~\citep{sutton2007role,janjua2024gvfs}. \looseness=-1

Much of the progress in RL has been driven by empirical advances, with experimental platforms often shaping algorithmic innovation and research directions. Consequently, evaluation platforms play a central role in determining what the field studies and how progress is measured~\cite[e.g.,][]{todorov2012mujoco,bellemare2013arcade,beatie2016deepmindlab}. In continual RL, most evaluation platforms are adaptations of traditional RL benchmarks, modified to introduce non-stationarity and thereby reflect the idea that the world is bigger than the agent. This is typically done by artificially switching tasks during interaction~\citep{powers2022cora,abbas2023loss,anand2023prediction,tomilin2023coom}, or, less commonly, by designing new environments specifically for continual RL that abandon episodic structure altogether~\citep{platanios2020jelly}.

However, both approaches have limitations. Reusing existing environments—particularly complex simulators—is appealing, yet introducing non-stationarity via hand-crafted task switches that are disconnected from the agent’s behaviour primarily models exogenous, abrupt regime shifts. Although such shifts occur in practice, many real-world environments evolve gradually and continuously through interaction. This endogenous, interaction-driven non-stationarity remains comparatively underexplored in continual RL benchmarks. Designing environments specifically for continual RL is promising, but existing instantiations often remain limited in complexity or scope. Our work takes a complementary direction: we retain the non-episodic structure of purpose-built environments while introducing rich, evolving stochastic dynamics, continuous control, partial observability, resource-driven competition, and progressively increasing behavioural demands within a spatially structured world.

\begin{figure}
\begin{center}
  \includegraphics[width=\textwidth]{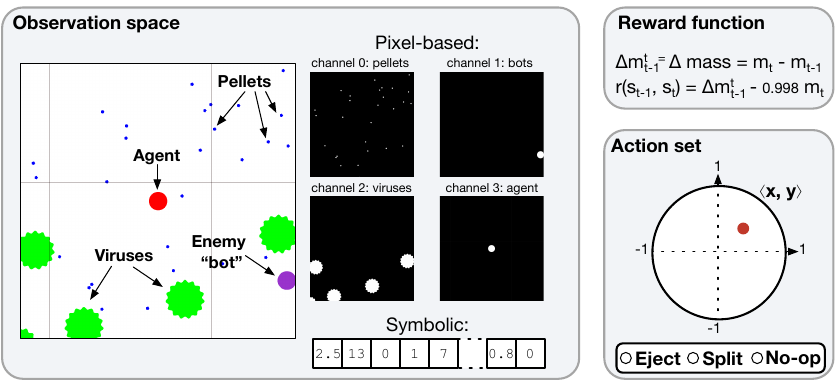}
\caption{\textbf{Agent–environment interface and main entities in \texttt{AgarCL}}. The agent receives either pixel-based or symbolic observations. The pixel-based input consists of four channels representing pellets, bots, viruses, and the agent itself, while the symbolic alternative provides pre-processed features such as distances to nearby entities. The reward is defined as the change in the agent’s mass between consecutive time steps. The action space is hybrid: at each step, the agent selects a continuous $\langle x, y \rangle$ coordinate (analogous to mouse control) and chooses whether to move, split, or eject mass. Section~\ref{AgarCL_Learning_Environment} provides further details on the entities, dynamics, and action definitions.
}
  \label{fig:agar_entities}
  \end{center}
\end{figure}

Concretely, we introduce a new evaluation platform based on the game Agar.io (Figure~\ref{fig:agar_entities}), which we call \texttt{AgarCL}. The platform was designed explicitly to support continual RL research. In this environment, the agent controls circular cells within a bounded, Petri dish–like arena and must accumulate mass in an ever-changing ecosystem of competing entities. Observations are high-dimensional and pixel-based,\footnote{Alternative symbolic observations, providing backward compatibility with related platforms~\citep{zhang2023gobigger}, are also supported; see Section~\ref{AgarCL_Learning_Environment}.} and the action space combines continuous navigation—parameterized as $\langle x, y \rangle$ movement—with discrete affordances such as splitting or ejecting mass. The reward is defined as the change in the agent’s mass between consecutive time steps, so effective behaviour corresponds to sustained growth over an indefinite horizon.

The dynamics of Agar.io make it particularly well suited to studying continual RL. The agent operates in an ecosystem populated by other agents,\footnote{In \texttt{AgarCL}, the behaviour of other agents is governed by hand-coded policies. Extending the platform to support learning across multiple agents is a promising direction for future work.} competing for mass in a bounded arena. Agents grow by collecting stationary food pellets or absorbing smaller cells, and to do so, they have to simultaneously avoid larger threats. At every time step, mass decays at a rate that increases with size, creating sustained pressure to act. Crucially, beyond its potential never-ending nature and the complexity induced by
other agents’ behaviours, the environment’s dynamics change as a function of the agent’s mass. Larger agents move more slowly, and the observation space progressively zooms out to keep the agent’s body visible, altering the scale and distribution of perceptual input. As a result, both the consequences of actions and the observation stream evolve continuously with the agent’s state. Although abrupt events also occur—such as absorbing another cell or splitting due to strategic choice or viral interaction—the dominant form of non-stationarity is smooth and interaction-driven. Figure~\ref{fig:agar_dynamics} and a YouTube video\footnote{https://www.youtube.com/watch?v=CGpvzHIqFLA.} depict some of the environment dynamics supported by \texttt{AgarCL}.

In this paper, we introduce \texttt{AgarCL} as both an evaluation platform and a challenge problem for continual RL. We evaluate three widely used RL algorithms—DQN~\citep{DBLP:journals/nature/MnihKSRVBGRFOPB15}, PPO~\citep{PPO}, and SAC~\citep{SAC}—as well as three approaches explicitly designed for continual learning: Shrink and Perturb~\citep{ash2020warm}, ReDo~\citep{sokar2023dormantneuronphenomenondeep}, and Continual Backpropagation~\citep{dohare2024loss}. Despite their strong performance in many established benchmarks, none of these methods achieves sustained competence in the full game. To better understand these failures, we introduce a suite of mini-games that isolate specific challenges embedded in the full environment, including non-stationarity, exploration, and long-horizon credit assignment. These controlled experiments reveal that the difficulties in \texttt{AgarCL} cannot be attributed solely to stability–plasticity trade-offs. Finally, our empirical study exposes persistent challenges in continual RL research, particularly regarding evaluation methodology and hyperparameter sensitivity, suggesting that progress in the field depends not only on new algorithms but also on more robust benchmarking practices.

\section{Background}
\label{Background}

We adopt the standard RL formalism to describe the sequential decision-making problem posed by \texttt{AgarCL}. Interactions occur at discrete time steps. The agent begins in a state $S_0 \sim \mu$, where $\mu$ is a start-state distribution over $\mathscr{S}$, but only observes $\omega_0 \in \Omega$, generated by an observation function $\Phi: \mathscr{S} \rightarrow \Omega$. At each time step $t$, the agent selects an action $A_t \in \mathscr{A}$ according to a (possibly stochastic) policy $\pi$ that depends on its observation history, and receives a reward $R_{t+1} \in \mathbb{R}$ and a new observation $\omega_{t+1}$. The agent's goal is to maximize some variant of the expected return, $G_t$, that is, of the expected total sum of rewards it receives. The most common variant is the \emph{discounted} return, $G_t^\gamma$, in which later rewards are discounted by $0 \leq \gamma < 1$, such that $G_t^\gamma \doteq \sum_{k=0}^\infty \gamma^k R_{t+k+1}$. Alternatively, one may consider the \emph{average-reward} setting: $G^{\bar{r}}_t \doteq \sum_{k=0}^\infty \big(R_{t+k+1} - r(\pi)\big)$, where $r(\pi) \doteq \lim_{h \rightarrow \infty} \frac{1}{h} \sum_{t=1}^{h}\mathbb{E}[R_t | \omega_0, A_{0:t-1} \sim \pi]$, with $\omega_0$ denoting the initial observation.

The environment evolves according to a transition function $p: \mathscr{S} \times \mathscr{A} \rightarrow \mathcal{P}(\mathscr{S})$, where $\mathcal{P}(\mathscr{S})$ denotes the set of probability measures over the state space. In MDPs~\citep{puterman2014markov}, observations coincide with states. However, \texttt{AgarCL} is partially observable and more appropriately modelled as a POMDP~\citep{kaelbling1998planning}, with complex dynamics hidden from the agent. This partial observability can induce apparent non-stationarity from the agent’s perspective, even when the underlying transition dynamics are stationary. As discussed above, we adopt the view that continual RL is beneficial in such settings because continual adaptation may outperform any fixed policy~\citep[see][]{sutton2007role,janjua2024gvfs}. This perspective allows us to remain agnostic to restrictive assumptions about agent capacity when invoking the “big world” hypothesis~\citep{javed2024big}.

In this work, we evaluate performance using the undiscounted return, while training algorithms derived from the discounted formulation. Although the average-reward objective is arguably more natural for continuing tasks, we leverage the substantially more mature deep RL literature developed for the discounted setting. This choice mirrors common practice. For example, in Atari 2600 games, agents are evaluated on accumulated scores while trained to optimize discounted return~\citep{bellemare2013arcade,machado2018revisiting}. Moreover, for every problem, there exists a critical discount factor such that any solution using a higher value will maximize the average reward~\citep{blackwell1962discrete}. \looseness=-1


\section{\texttt{AgarCL}: Agar.io for Continual Reinforcement Learning}
\label{AgarCL_Learning_Environment}

\begin{figure}[t]
\begin{center}
  \includegraphics[width=\textwidth]{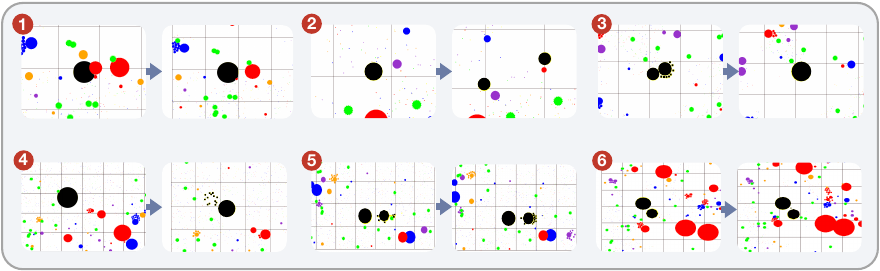}
  \caption{
  \textbf{Environment dynamics and actions.}
\textcircled{1}~The agent (black) consumes smaller cells to gain mass.
\textcircled{2}~The \textsc{Split} action divides each controlled cell in half and propels one of the resulting cells in a chosen $\langle x, y \rangle$ direction, enabling slower, larger cells to reach faster opponents. Cell speed decreases as mass increases.
\textcircled{3}~Separated cells can later merge when brought sufficiently close together.
\textcircled{4}~Depending on its mass, a collision with a virus either fragments the agent into multiple cells or
\textcircled{5}~allows the agent to absorb the virus.
\textcircled{6}~The agent may also \textsc{eject} mass in a chosen direction. Ejected mass can be consumed by any cell and, if sufficient mass is fed into a virus, causes it to spawn a new virus while propelling the original, enabling smaller agents to attack larger ones. \looseness=-1} 
  \label{fig:agar_dynamics}
  \end{center}
\end{figure}

We begin by discussing our main contribution: Agar.io for Continual RL (\texttt{AgarCL}), a research environment for sequential decision-making in which the dynamics evolve continuously as a function of the agent’s state and its interactions with other agents. Unlike benchmarks that rely on explicit task switches, \texttt{AgarCL} induces endogenous, state-dependent changes in both control and observation, while other agents in the environment further shape the experienced dynamics. Together, these factors result in a setting where the effective environment shifts over time without externally imposed task boundaries. This aligns with the ``big world'' hypothesis~\citep{javed2024big,lewandowski2025world}, as the agent operates in a world whose complexity is not exhaustively captured within a fixed training horizon and where ongoing adaptation is expected to remain beneficial.

\subsection{The Game: Agar.io}

\texttt{AgarCL} is based on the multiplayer online game Agar.io, in which each player controls one or more circular cells in a shared two-dimensional arena. The setting resembles a Petri dish populated by interacting cells, food pellets, and viruses. Players increase their mass by absorbing smaller entities, including stationary pellets and other players’ cells, while avoiding larger opponents. Viruses introduce additional strategic structure: depending on a cell’s size, colliding with a virus can either fragment it into multiple smaller cells or provide temporary protection against larger threats (see Figure~\ref{fig:agar_entities}). Players may also voluntarily split their cells for tactical purposes, allowing simultaneous control of multiple bodies. Although the underlying rules are simple, complex dynamics emerge from these local interactions. Figure~\ref{fig:agar_dynamics} illustrates several of these behaviours.\looseness=-1

The arena contains three types of entities: \emph{pellets}, \emph{cells}, and \emph{viruses}. Pellets are randomly scattered, static, have fixed size, and grant one unit of mass when consumed. Cells (including both the learning agent and bots) can consume pellets, viruses, and smaller cells; when a cell absorbs another cell, it gains its entire mass. Viruses have a fixed mass of $100$, and depending on the size of the colliding cell, a virus may either be absorbed or cause the cell to fragment into multiple smaller cells. \looseness=-1

Players begin as a small cell with mass 25. If a player’s cell is consumed by a larger cell, it is eliminated and subsequently respawns with the same initial mass. Although this repeated elimination might suggest an episodic structure, \texttt{AgarCL} is fundamentally a continual task: successful agents could survive forever, and, crucially, the environment does not reset upon elimination. The cell that absorbed the player retains its newly acquired mass, so the consequences of actions taken in one ``life'' persist into subsequent ones. Our current implementation focuses on the single-agent setting, where other players, called \emph{bots}, follow heuristic-based policies. This design isolates the continual adaptation problem while maintaining rich interaction dynamics. For researchers interested in fully multi-agent variants of the game, we refer to the \textsc{GoBigger} platform~\citep{zhang2023gobigger}. \looseness=-1

\begin{figure}
\begin{center}
  \includegraphics[width=\textwidth]{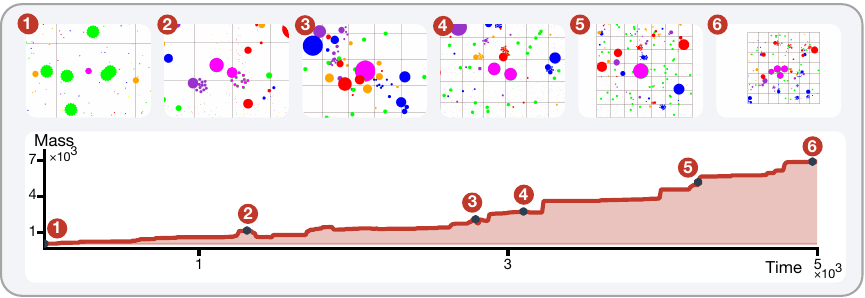}
  \caption{\textbf{The first 5,000 time steps of an expert trajectory showing game progression.} The agent (pink) steadily gains mass: \textcircled{1}~starting small, it eventually passes over a virus; \textcircled{2}~splits into smaller cells; \textcircled{3}~grows sufficiently large to consume viruses and other agents; \textcircled{4}~splits to attack, increasing its speed. As mass increases, the field of view expands to keep all controlled cells visible. \textcircled{5}~The agent ultimately surpasses all opponents in mass and \textcircled{6}~can see the entire arena. 
  \looseness=-1}
  \label{fig:agar_progression}
  \end{center}
\end{figure}

Several mechanisms contribute to the game’s evolving dynamics. New pellets are generated every 600 ticks (10 seconds) whenever fewer than 500 pellets are present in the arena, and viruses are generated at each time step if their total number falls below 10. At the same time, each cell loses 0.2\% of its mass per second, thereby imposing a constant pressure to grow. As mass increases, mobility decreases: a cell’s speed $v$ is determined by $v = \nicefrac{100}{\mathrm{mass}^{0.439}}$, resulting in a slower movement for larger cells. Consequently, smaller players must evade larger opponents, avoid being trapped near boundaries, or strategically exploit viruses for protection. Players can feed viruses mass; once sufficiently large—typically after seven feedings—a virus splits and can be directed toward nearby cells, potentially fragmenting larger opponents. A player’s field of view also expands with its mass to ensure that all controlled cells remain visible.\footnote{The original game contains additional rules. For example, a player cannot split beyond 14 cells, and consuming multiple viruses in succession activates a penalty that increases mass decay; this penalty persists across respawns. We include these rules to faithfully replicate the original game mechanics and minimize experimenter-induced bias, as Agar.io was not designed for AI research, and we wanted to preserve that as much as possible.} Figure~\ref{fig:agar_progression} illustrates a typical progression of play.

These dynamics create persistent trade-offs between short-term gains and long-term survival. For example, a common tactic is to split a large cell to increase the likelihood of consuming smaller opponents that would otherwise remain out of reach due to reduced mobility. However, this aggressive move introduces significant risks as the resulting smaller cells can be consumed by larger opponents. Such decisions require balancing offensive opportunities against defensive stability, with consequences that unfold over extended time scales. Because no single strategy is uniformly effective across mass regimes or local configurations of opponents, agents must continually adjust their behaviour in response to both their own evolving state and the surrounding ecosystem.

\subsection{Agent-Environment Interface}

Having specified the underlying dynamics, we now formalize the interaction protocol between the agent and the environment. We adopt the standard RL formulation in which interaction proceeds through discrete time steps via actions, observations, and rewards. We detail the action space, observation space, and reward structure below and summarize the key entities in Figure~\ref{fig:agar_entities}. \looseness=-1

\paragraph{Reward Function.} To keep the rewards bounded, we define the reward at time step $t$ as the change in the agent’s mass between consecutive time steps:
\begin{equation}
R_t = m_t - m_{t-1}.
\label{eq_diff}
\end{equation}
Importantly, the problem is continuing (non-episodic), so the agent’s death does not terminate the game. Upon respawning, the agent receives a reward equal to its death mass less its initial mass. This choice discourages death as a cost-free reset when the agent is bigger than its initial mass. Additionally, each cell loses 0.2\% of its mass per second, which we do not represent in Equation~\ref{eq_diff}.

\looseness=-1

\paragraph{Observation Space.}

How the agent perceives the environment critically determines the difficulty and structure of the learning problem. We primarily consider a pixel-based, top-down rendering of the environment, requiring the agent to operate over a high-dimensional, partially observable, and non-stationary visual stream. Figures throughout the paper illustrate examples of the agent's observations, with Figure~\ref{fig:agar_progression} highlighting the evolution of the observation stream.

Variants of Agar.io have previously been explored in RL research~\citep{wiehe2018sampled,anso2019deep,zhang2023gobigger}; however, those works rely on structured grid-like or symbolic state representations. To the best of our knowledge, a high-dimensional pixel-based formulation has not been studied. While our experiments focus on the pixel-based setting, \texttt{AgarCL} also supports more symbolic observation variants for controlled comparisons. We describe both below.

{\emph{Pixel-Based Observation:}} The observation at time step $t$ is represented as a tensor 
$\mathbf{O}_t \in \mathbb{R}^{N \times N \times 4}$,
where $N \times N$ denotes the spatial resolution (default $N = 128$). The four channels encode pellets, viruses, enemy cells, and the agent (including gridlines), respectively. Figure~\ref{fig:agar_entities} provides an example rendering (gridlines are faint and may not be visible).

{\emph{Grid-Like Observation:}}
We additionally adapt the observation design from GoBigger~\citep{zhang2023gobigger} to the single-agent setting. This representation consists of a global state and a player state. The global state includes coarse environmental information such as map size and elapsed time steps. The player state captures agent-specific information, including its field of view, visible entities within that region, current score, and available actions. The overlap field is critical because it encodes nearby pellets, viruses, and cells together with their positions, velocities, and associated attributes.


\paragraph{Action Space.}

The action space is hybrid. At each time step, the agent selects both a continuous control vector and a discrete action. The continuous component corresponds to a cursor location $\langle x, y \rangle \in [-1,1]^2$, which determines the direction of movement for all controlled cells. This design mimics human play, in which movement is specified by pointing to a location in the arena.

Simultaneously, the agent might choose to select one discrete action from \{\textsc{Split}, \textsc{Eject}\}. Not selecting any actions applies only the continuous cursor control. The \textsc{Split} action divides each eligible cell into two equal-mass cells, provided the resulting cells have mass at least 25 (otherwise the action has no effect). One of the resulting cells is propelled toward the cursor direction with substantial momentum. After splitting, multiple cells must be controlled jointly through the same cursor input. The \textsc{Eject} action expels a small mass unit from each cell toward the cursor direction. Ejected mass can be consumed by cells or viruses. All selected actions are executed for four consecutive environment steps before a new action is sampled. The observation returned to the agent corresponds to the state after these four steps (i.e., a frame skip of 4).

Finally, the environment includes stochasticity: Gaussian noise $\mathcal{N}(0,1)$ is added to the continuous control signal before execution, preventing perfectly deterministic actuation.




\paragraph{Simulation Speed.}
We report performance in terms of the interquartile mean (IQM) over ten independent trials. In our experiments, a random Python agent processes 2,016 frames per second with a frame skip of 1. With the default frame skip of 4, the agent receives 1,163 observations per second, corresponding to 4,652 environment frames and 1,163 agent decisions per second.  This throughput is substantially higher than prior Agar.io-based benchmarks such as \textsc{GoBigger}. The performance gains result from several implementation choices, including EGL-based rendering, optimized collision detection, efficient observation generation, and careful memory management.

\paragraph{Technical Details.} Appendix~\ref{sec:release} provides implementation details regarding the release of \texttt{AgarCL}, including its software architecture, agent interface, configuration options, and performance.

\section{Experiments and Results}
\label{sec:experiments}

To evaluate the learning capabilities of agents in \texttt{AgarCL}, we begin with standard deep reinforcement learning baselines: DQN~\citep{DBLP:journals/nature/MnihKSRVBGRFOPB15}, PPO~\citep{PPO}, and SAC~\citep{SAC}.\footnote{Our implementations are available in \texttt{https://github.com/machado-research/AgarCL-benchmark}.}
 These algorithms span value-based and policy-gradient methods, and together represent widely used approaches for high-dimensional control. By starting with these methods, we assess whether existing deep RL techniques are sufficient to cope with the non-stationary, partially observable, and multi-agent dynamics induced by \texttt{AgarCL}.

All agents process $128 \times 128 \times 4$ pixel observations through a shared convolutional encoder. The encoder consists of three conv. layers with kernel sizes $8 \times 8$, $4 \times 4$, and $3 \times 3$, and strides 4, 2, and 1, respectively. Each layer is followed by layer normalization~\citep{ba2016layer} and a ReLU activation. The resulting $32 \times 12 \times 12$ feature map is flattened and passed through a fully connected layer to obtain an embedding. This embedding is then used to predict both discrete and continuous actions, enabling hybrid control. Appendix~\ref{appendix:algorithms} details the algorithm-specific adaptations required for this setting. \looseness=-1

\subsection{Benchmark Results for Continual RL in \texttt{AgarCL}}~\label{sec:benchmark_full_game}

The full game constitutes the reference and default configuration of \texttt{AgarCL}. We intentionally evaluate agents in this setting first, as it combines all the key challenges posed by \texttt{AgarCL}. The arena size is $350 \times 350$ and contains $10$ viruses, $8$ bots, and $500$ pellets that regenerate every $600$ ticks. We train DQN, PPO, and SAC for $160$ million environment frames, averaging results over $10$ independent runs. Training is computationally demanding: running SAC in this setting, for example, requires more than seven days on a single machine.

\begin{wrapfigure}[13]{r}{0.4\textwidth}
  \centering
  \includegraphics[width=\linewidth]{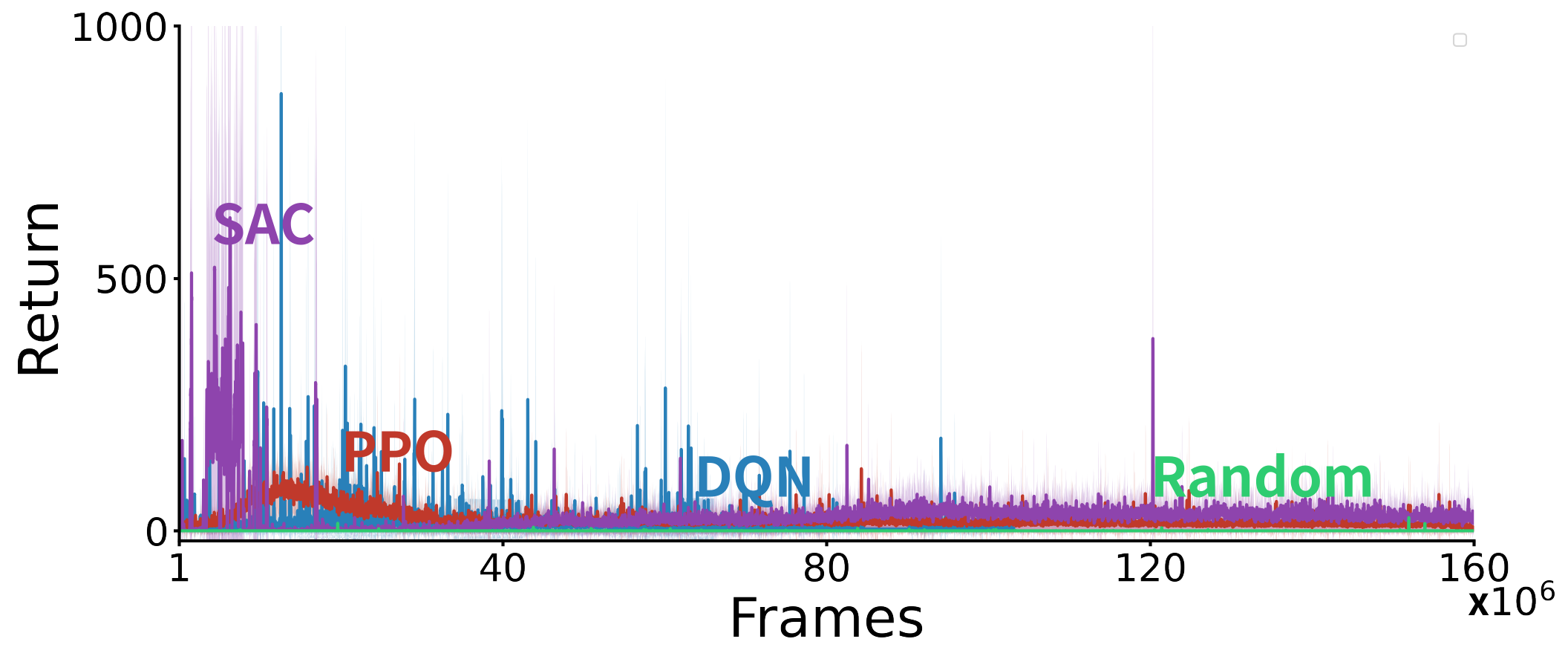}
    \caption{\textbf{Performance of established deep RL algorithms in the full game of \texttt{AgarCL}.} Each curve is the average over 10 seeds, with a moving average computed over a window of 100 steps.}
  \label{fig:full-game}
\end{wrapfigure}

The results highlight the difficulty of \texttt{AgarCL} for standard deep RL methods. Across all algorithms, we observe a consistent failure to learn an effective policy in the full-game configuration. Figure~\ref{fig:full-game} shows the learning curves, and exact numerical results are provided in Table~\ref{table_results} (Appendix~\ref{appendix:Table_Results}). These outcomes indicate that traditional approaches struggle to address the challenges surfaced by \texttt{AgarCL}, which are central to continual RL. We analyze these challenges in more detail in the following sections. \looseness=-1

Hyperparameter tuning is a central yet often overlooked challenge in continual RL. In continual settings, standard tuning strategies, such as optimizing performance over a fixed horizon, are inherently problematic~\citep{mesbahi2024tuning}. If an agent is expected to operate indefinitely, what horizon should guide tuning? Selecting a shorter interval risks overfitting to that duration, contradicting continual learning. This issue is not merely theoretical: e.g., hyperparameters that perform well on 400k frames in Atari 2600 games often differ substantially from those used when training for 200M frames. Moreover, RL performance is frequently highly sensitive to small hyperparameter changes (see Appendix~\ref{appendix:hyper_sensitivity}). \looseness=-1

Because tuning specifically for the 160M-frame horizon of the full game would implicitly bias the evaluation toward that timescale, we avoid horizon-specific tuning. At the same time, hyperparameters must be chosen. We therefore reuse the best hyperparameters identified for \texttt{AgarCL}'s continual \textsc{Mini-Game~4}, the setting most similar to the full game but considerably shorter. This pragmatic choice reflects a broader open question in continual RL: how should hyperparameters be selected when no natural training horizon exists?



\subsection{\texttt{AgarCL} as a Continual Reinforcement Learning Testbed}
\label{AgarCL_ExpV2}

We have not yet provided direct evidence that \texttt{AgarCL} meaningfully captures the challenges of continual RL. We do so here by examining whether continual learning confers an advantage. Specifically, we compare agents whose policies are fixed after training with agents that learn continually.

\begin{wrapfigure}[15]{r}{0.48\textwidth}
    \centering
    \includegraphics[width=0.46\textwidth]{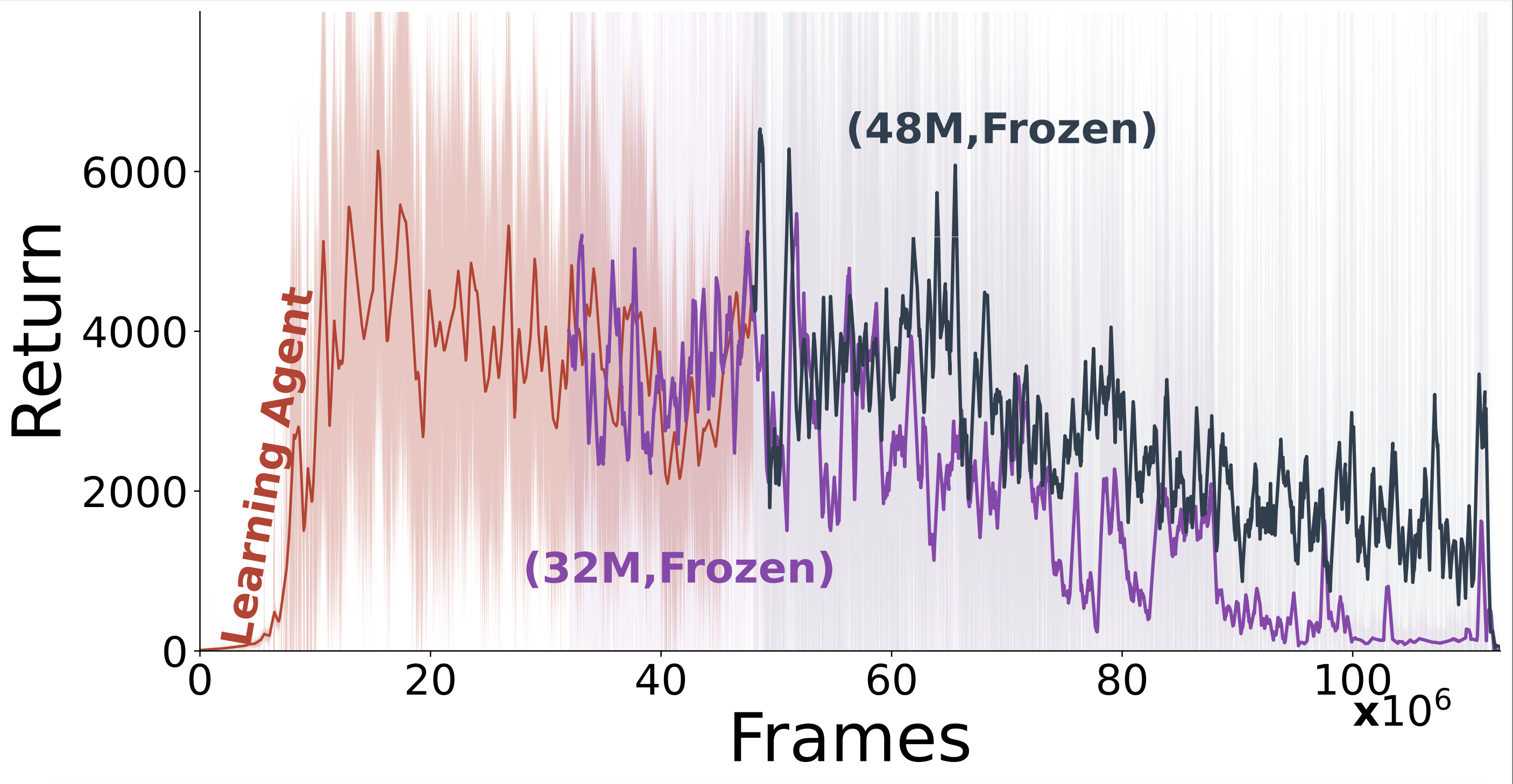}
    \caption{\textbf{Performance of fixed-policy agents initialized from checkpoints at 32M and 48M steps.} We report the moving average over 10 random seeds with a window size of 1000 steps. \looseness=-1}
    \label{fig:PPO_performance_different_continual_fixed}
\end{wrapfigure}

Demonstrating this effect in the full-game setting is difficult, as the algorithms evaluated in Section~\ref{sec:benchmark_full_game} did not learn a reasonable policy. Thus, we consider an easier configuration in which pellets and viruses regenerate every 120 ticks (instead of 600), and pellet density is increased to 1024 (instead of 500). We focus on PPO here because it consistently outperformed the other baselines across multiple configurations, as discussed in the next sections. Performance across different pellet densities is reported in Appendix~\ref{appendix: PPO_Performance_sparse_env}. \looseness=-1

Having identified a setting in which PPO learns a reasonable policy, we evaluate the robustness of that policy when learning is halted. Specifically, we freeze policies trained for $32$ million and $48$ million steps and evaluate them in the configuration with $1024$ pellets. As shown in Figure~\ref{fig:PPO_performance_different_continual_fixed}, although the frozen agents initially perform competitively, their performance eventually collapses. This behavior indicates that policies learned at a fixed point in time fail to remain effective as the environment evolves, supporting the use of \texttt{AgarCL} as a testbed for continual adaptation.

It is also informative to consider a configuration where this collapse does not occur. When reducing the number of bots from eight to four, PPO learns a relatively stable policy that collects pellets and reliably avoids opponents (see Appendix~\ref{app:continual_learning}). In this simpler setting, freezing the policy does not lead to the same degradation in performance. This contrast highlights the role of environmental complexity in inducing continual adaptation and underscores how specific design elements of \texttt{AgarCL} contribute to its suitability as a continual RL evaluation framework.

\subsection{Existing Continual RL algorithms in \texttt{AgarCL}}\label{app:continual_backprop}

While the primary goal of this paper is not to benchmark continual RL methods in \texttt{AgarCL}, it is natural to ask whether existing approaches transfer to this setting.

Many established continual learning methods rely on explicit task boundaries, which are not available in \texttt{AgarCL}. Representative examples include EWC~\citep{Kirkpatrick2017overcoming}, MAS~\citep{Aljundi2018memory}, and LwF~\citep{Li2016Learning}. EWC and MAS incorporate task identity into regularization constraints on parameter updates, while LwF uses task structure to define distillation losses that preserve performance on earlier tasks.

\begin{wrapfigure}[14]{r}{0.48\textwidth}
    \centering
    \includegraphics[width=0.46\textwidth]{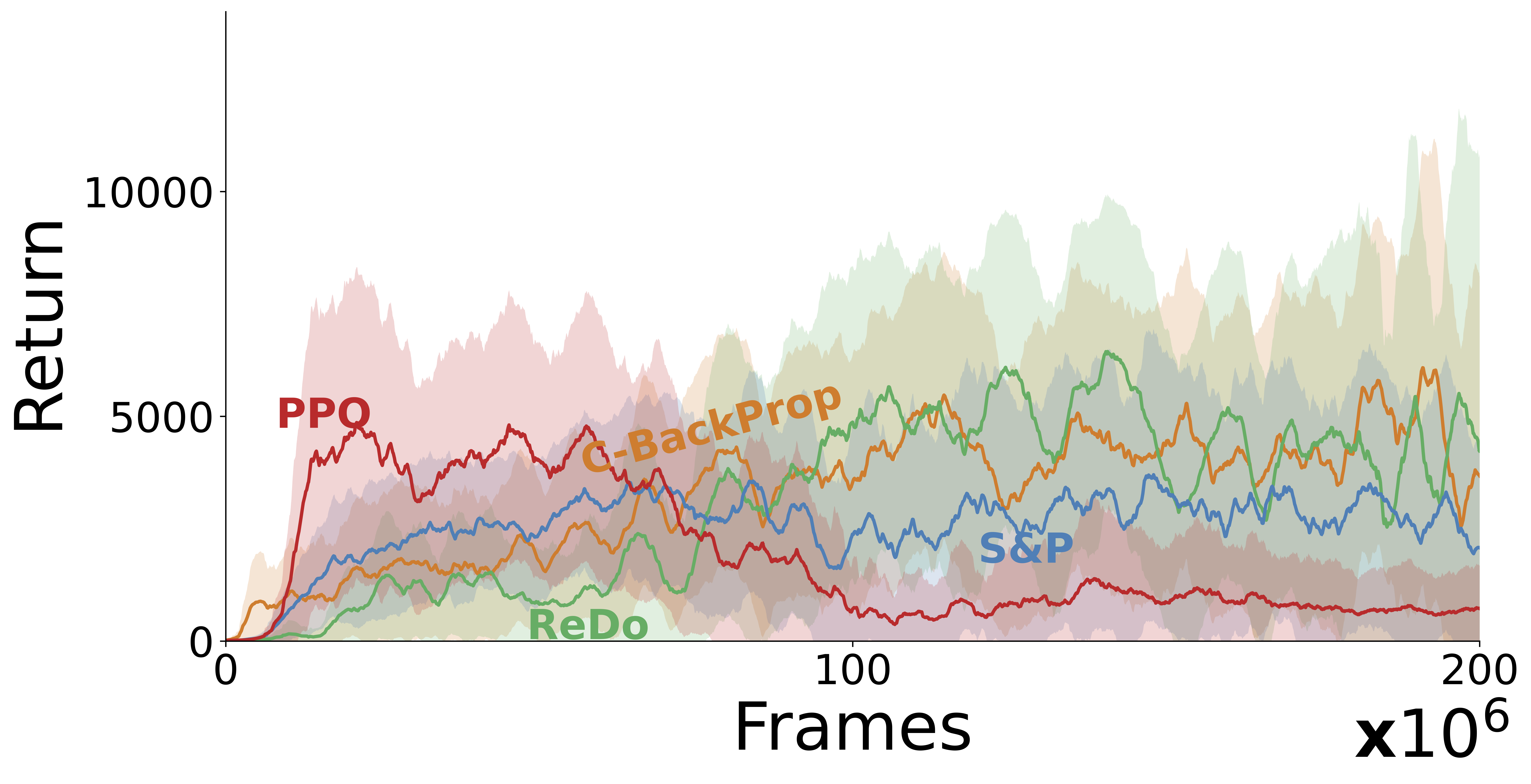}
\caption{\textbf{Performance of different continual learning approaches.}~We report the moving average over 10 runs with a window size of 10,000 steps.\looseness-1 }
\label{fig:PPO_Performance_CL}
\end{wrapfigure}

We therefore focus on task-agnostic methods that can be applied to \texttt{AgarCL} without structural modifications. These approaches primarily aim to preserve plasticity during learning. Concretely, we augment PPO with Shrink and Perturb~\citep{ash2020warm}, ReDo~\citep{sokar2023dormantneuronphenomenondeep}, and Continual Backpropagation~\citep{dohare2024loss}, and evaluate them in the easier setting introduced in Section~\ref{AgarCL_ExpV2}. \footnote{Other replay-based continual learning approaches, such as CLEAR~\citep{Rolnick2019Experience}, could also be considered. CLEAR leverages past experience from a replay buffer to mitigate forgetting. In this work, however, our goal is to analyze how environment design influences continual adaptation rather than to exhaustively benchmark all continual RL baselines.} As shown in Figure~\ref{fig:PPO_Performance_CL}, the performance of the baseline, PPO, starts to drop before 100 million frames, while we do not observe this behaviour with PPO with techniques designed to mitigate loss of plasticity. Hyperparameters and numerical results are reported in Appendix~\ref{app:continual_learning}. \looseness=-1

We did not perform any additional investigation to determine the extent to which PPO's performance degradation is attributable to loss of plasticity. Nevertheless, the observation that equipping PPO with techniques designed to mitigate loss of plasticity tend to achieve higher average performance, albeit with substantial variance, suggests that loss of plasticity may contribute to the observed performance degradation.
More importantly, we observe that none of the methods achieve strong performance, and we saw little evidence of learning in the more challenging configuration with 500 pellets and 600-tick regeneration. These findings suggest that while stability–plasticity may be one contributing factor in \texttt{AgarCL}, the primary bottlenecks likely lie elsewhere, including exploration and long-horizon credit assignment.



\subsection{Validating \texttt{AgarCL} through Mini-Games}

As shown above, \texttt{AgarCL} presents a highly challenging learning problem. Beyond requiring sustained online adaptation, it exposes additional difficulties central to continual RL, including exploration without resets, long-horizon credit assignment, and stable representation learning under a continuously evolving observation stream.

To better understand these challenges, we evaluate agents on a set of mini-games designed within Agar.io. Each mini-game isolates specific aspects of the full environment, such as non-stationarity and its non-episodic dynamics, allowing us to analyze their impact in a more controlled manner.

Importantly, we tuned each algorithm's hyperparameters in every mini-game. We did so over 3 random seeds, selecting the configuration that maximized the mean return across the final 100 episodes of a 20-million-frame run. To avoid maximization bias, we evaluated this configuration with 10 new trials. Details on the hyperparameter ranges and selection criteria are provided in Appendix~\ref{appendix:training-details}. 
\looseness=-1

\textbf{Non-Stationarity through Mass Variation.} We begin with a set of progressively challenging pellet-collection mini-games. These environments exclude bots and viruses, isolating the effect of mass dynamics. For simplicity, this first group is \emph{episodic}, with fixed episode lengths of 500 (first three mini-games) and 3000 (last three mini-games) time steps, and a single start state.

These mini-games were designed to isolate three aspects of the full game: (i) short-term exploration, (ii) mass decay, and (iii) the dynamics induced by large mass (slower movement but access to splitting). To simplify exploration, delayed credit assignment, and partial observability, \textsc{Mini-Game~1} consists of collecting pellets arranged in a square path. The agent starts with mass 25 and experiences no mass decay. Although this is the simplest setting, it still exhibits non-stationarity: as the agent gains mass, it becomes larger and slower, yet the optimal behavior is simply to follow the dense trail of pellets. \textsc{Mini-Game~2} extends this setup by introducing mass decay, where the loss per step depends on the agent’s current mass. \textsc{Mini-Game~3} further increases the starting mass to 1000, making the agent substantially slower, increasing decay, and enabling splitting. \textsc{Mini-Games~4–6} mirror the first three settings, except that pellets are randomly scattered rather than arranged along a dense path. In these variants, credit assignment is more delayed, exploration requirements change, and partial observability plays a larger role (though episodic resets still simplify the task). Screenshots of all mini-games are provided in Appendix~\ref{sec:mini-games}.

Figure~\ref{fig:pellets_mini_games} reports performance across all mini-games; numerical results are provided in Appendix~\ref{appendix:Table_Results}. In the simplest setting, \textsc{Mini-Game~1}, all algorithms achieve performance comparable to that of a human player. Introducing mass decay substantially increases difficulty, even in the structured square-path task, reducing performance to roughly half of the human baseline. Starting the agent with a larger mass further amplifies this effect. These trends persist when pellets are uniformly distributed across the arena, although the task becomes markedly more challenging. In these variants, and particularly once mass decay is introduced, only PPO is able to learn a non-trivial policy.

\begin{figure}[t]
\begin{center}
\includegraphics[width=\textwidth]{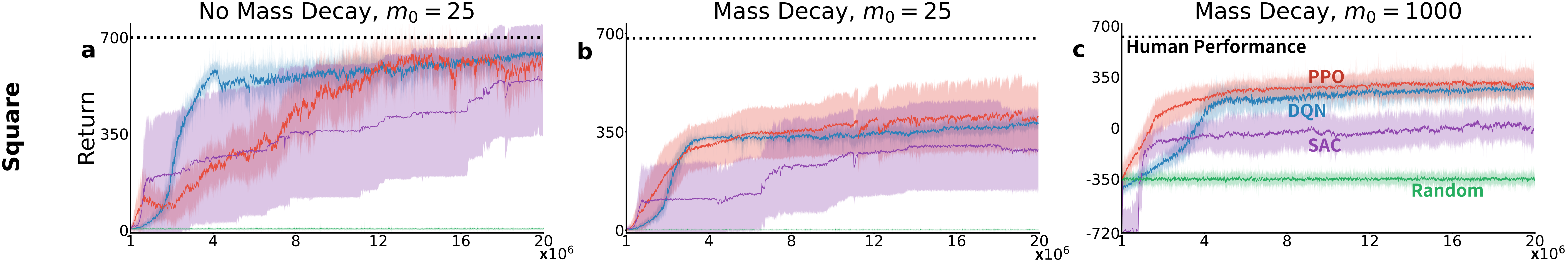}
\includegraphics[width=\textwidth]{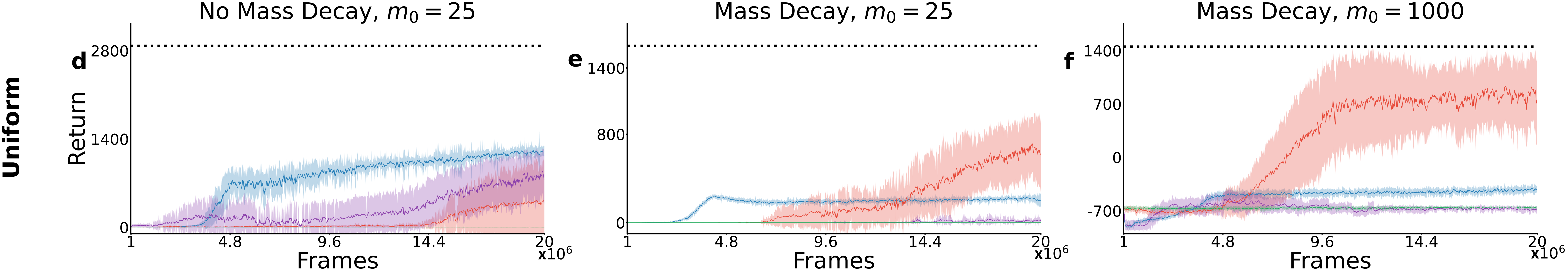}
\caption{\textbf{Performance of RL methods on \emph{episodic} pellet-collection mini-games.} Panels \textcircled{a}–\textcircled{c} show results for the square-path tasks (\textsc{Mini-Games~1–3}), while \textcircled{d}–\textcircled{f} correspond to the randomly regenerated tasks (\textsc{Mini-Games~4–6}). Y-axis scales differ across panels. The dashed line indicates human performance, and the green line denotes the random policy. Shaded regions represent 95\% confidence intervals over 10 runs, computed using the t-distribution. \looseness=-1}
\label{fig:pellets_mini_games}
\end{center}
\end{figure}

\textbf{Continual Problems.} We now evaluate the mini-games above in the continual setting.\footnote{We use the word \emph{continual} deliberately to emphasize the non-stationary nature of the problem. \emph{Continuing} typically refers to stationary infinite-horizon problems, while \emph{continuous} denotes continuous action spaces.} We remove episodic resets and replenish all pellets every 600 frames. The maximum number of pellets is 500, regenerated randomly (restricted to the path in the square-path mini-games). As before, we re-tuned hyperparameters separately for each mini-game and algorithm.

These tasks proved substantially more challenging. Results are reported in Appendix~\ref{appendix:Table_Results} (Figure~\ref{fig:uniform_pellets_cont} and Table~\ref{table_results}). In the square-path variants, no agent succeeds. Exploration and partial observability play a central role: once the agent deviates from the path and loses sight of it, it struggles to recover, and without resets it must rediscover the path to continue learning. Interestingly, when pellets are uniformly distributed, performance improves relative to the square-path setting; possibly because pellets remain regularly observable. However, reducing the pellet count by half makes the problem considerably harder, and the baselines degrade sharply (Appendix~\ref{appendix: PPO_Performance_sparse_env}). As before, PPO is the most robust method among those evaluated. For \textsc{Mini-Game~4}, we truncate training once agents reach the maximum allowed mass; without mass decay they would otherwise grow unbounded.

Finally, one might expect architectures with explicit memory to perform better in this setting. We therefore evaluated PPO augmented with a GRU~\citep{cho2014properties}. We did not observe consistent improvements that justified the added complexity, and thus focus on simpler architectures in the main text. Detailed results are provided in Appendix~\ref{sec:experiments_rnns}.

\textbf{Interacting with Other Agents.} We also designed mini-games in which the agent shares the arena with a second bot governed by a fixed policy. We evaluated performance against several simple opponents. Across all cases, the learning agents failed to learn a policy that reliably collected enough pellets to outgrow and eventually absorb the opponent, even when the opposing bot did not actively chase them. These results further highlight the difficulty of exploration in this environment: even acquiring the seemingly basic skill of safely scaling before engaging proves non-trivial. Additional details are provided in Appendix~\ref{appendix:other_agents}.

\textbf{Interacting with Viruses.} Beyond pellet collection and direct competition, we designed a mini-game to assess whether agents could learn to use viruses strategically, splitting a larger bot and subsequently absorbing it. As in the previous settings, none of the evaluated agents learned the required sequence of actions, despite simplifying the task as much as possible. These results help explain why prior work in Agar-like environments has largely focused on pellet-collection tasks rather than other strategic interactions. Detailed results and analysis are provided in Appendix~\ref{sec:viruses_minigame}.

\textbf{Summary.} Taken together, these mini-game results illustrate the range and severity of challenges posed by \texttt{AgarCL}. Even when individual aspects of the environment are isolated—non-stationarity, long-term credit assignment, partial observability, or strategic interaction—existing methods struggle to achieve robust performance. The transition from episodic to continual variants further amplifies these difficulties, revealing sharp degradation once resets are removed. More complex interactions, such as competition and virus-based strategies, remain entirely unsolved. Collectively, these findings validate \texttt{AgarCL} as a demanding and realistic benchmark for continual reinforcement learning, where progress requires advances beyond incremental algorithmic refinements.


\section{Related Work}
\label{sec_RelatedWork}

\textsc{GoBigger}~\citep{zhang2023gobigger} is the closest existing platform to ours due to its Agar-style gameplay. However, it was designed to study collective behaviors in multi-agent RL, emphasizing coordination among \emph{teams} of agents. It does not target continual RL; all tasks are episodic. In contrast, \texttt{AgarCL} is explicitly designed for continual settings, supports additional data streams, and scales much better with increasing numbers of cells and pellets. Earlier work also explored Agar-style games as RL environments, primarily to benchmark deep RL algorithms on pellet-collection tasks~\citep{anso2019deep,wiehe2018sampled}, rather than to study non-stationarity or continual RL. \texttt{AgarCL} is implemented on top of the AgarLE \citep{AgarLE}, an incomplete implementation of Agar.io with an OpenAI Gym~\citep{brockman2016openai} interface.

Many platforms have been proposed for continual RL research. \texttt{AgarCL} complements these efforts by introducing a fundamentally different problem structure. In particular, unlike Switching ALE~\citep{abbas2023loss}, Continual World~\citep{wolczyk2021continual}, POET~\citep{POET}, MEAL~\citep{tomilin2025mealbenchmarkcontinualmultiagent}, Continual NavBench~\citep{kob2025continual}, and numerous MiniGrid variants~\citep{ChevalierBoisvert23minigrid}, \texttt{AgarCL} is not organized as a sequence of episodic tasks that periodically switch. Instead, non-stationarity emerges organically from the underlying dynamics of a persistent environment.

The primary platform explicitly supporting non-episodic continual RL is JellyBean World~\citep{platanios2020jelly}. However, it features comparatively simple dynamics, a discrete action space, and limited non-stationarity, which primarily stems from abrupt reward-function changes. Other alternatives include bespoke adaptations of MuJoCo~\citep{todorov2012mujoco} or IsaacGym~\citep{makoviychuk2021isaac} environments used in specific studies~\citep[e.g.,][]{feng2022factored,long2024learning}. These setups, however, were not released as standardized evaluation platforms. Beyond individual features, a key distinction is that \texttt{AgarCL} integrates reward, perception, and dynamics in a tightly coupled manner, producing non-stationarity that is directly linked to the agent’s state and behavior. Reproducing this form of smoothly evolving, interaction-driven non-stationarity within physics engines would require substantial additional design.

Finally, complex environments such as NetHack~\citep{nethack} and Minecraft~\citep{johnson2016malmo,Guss2019MineRL} can also serve as potential evaluation frameworks for continual RL under the premise that they are ``bigger than the agent''~\citep{javed2024big}. However, their richness introduces additional challenges that may obscure the specific difficulties of continual learning. NetHack, for example, relies heavily on language and prior knowledge, and existing approaches typically require substantial auxiliary machinery. Likewise, many successful Minecraft solutions rely on human demonstrations because exploration is difficult. In contrast, \texttt{AgarCL} aims to isolate continual learning challenges without relying on external supervision or domain-specific priors.


\section{Conclusion}

We introduced \texttt{AgarCL} as an evaluation platform for continual reinforcement learning. It captures key challenges, including partial observability and smooth, endogenous non-stationarity, while avoiding the abrupt task switches common in existing benchmarks. Through extensive experiments with DQN, PPO, and SAC across mini-games and the full environment, we demonstrated both the severity of the full problem and the utility of the mini-games for controlled analysis.

\texttt{AgarCL} incorporates features central to continual RL, including non-episodic interactions, smooth non-stationarity, high-dimensional observations, continuous actions, and a potentially unbounded horizon. Our results highlight the limitations of standard deep RL algorithms in such settings and underscore the gap between current methods and the demands of continual learning. Beyond introducing capabilities largely absent from prior frameworks, we show that even fixed policies fail to maintain stable performance in this evolving environment.

This work emphasizes the problem itself rather than proposing new solution methods. Given both scope and space constraints, we did not systematically evaluate algorithms specifically designed for continual RL. A practical limitation is the computational cost: although \texttt{AgarCL} was engineered for efficiency, continual evaluation at the required time scales remains resource-intensive, and experimental cycles can still be lengthy.

\section*{Acknowledgments}

We would like to thank 	Jan Prus-Czarnecki and Anton Wiehe for early discussions, Alex Lewandowski and Anna Hakhverdyan for advice throughout the project, and Brett Daley and Prabhat Nagarajan for feedback on an earlier version of the manuscript. We would also like to thank Jon Deaton and the other contributors of the AgarLE GitHub repository ---\texttt{github.com/jondeaton/AgarLE}. Kateryna contributed to this project while interning at NVIDIA and working in Prof. Machado's lab.

The research is supported in part by the Natural Sciences and Engineering Research Council of Canada (NSERC) and the Canada CIFAR AI Chair Program. This research was enabled in part by the computational resources provided by the Alberta Machine Intelligence Institute (Amii), the Digital Research Alliance of Canada, and the University of Alberta Google Cloud Incubator.

\bibliography{main}

@inproceedings{
feng2022factored,
title={Factored Adaptation for Non-Stationary Reinforcement Learning},
author={Fan Feng and Biwei Huang and Kun Zhang and Sara Magliacane},
booktitle={Neural Information Processing Systems (NeurIPS)},
year={2022},
}

@inproceedings{
makoviychuk2021isaac,
title={{Isaac Gym: High} Performance {GPU} Based Physics Simulation For Robot Learning},
author={Viktor Makoviychuk and Lukasz Wawrzyniak and Yunrong Guo and Michelle Lu and Kier Storey and Miles Macklin and David Hoeller and Nikita Rudin and Arthur Allshire and Ankur Handa and Gavriel State},
 booktitle = {Neural Information Processing Systems (NeurIPS)},
 year={2021},
}

@inproceedings{kob2025continual,
  author       = {Anthony Kobanda and
                  Odalric{-}Ambrym Maillard and
                  R{\'{e}}my Portelas},
  title        = {A Continual Offline Reinforcement Learning Benchmark for Navigation
                  Tasks},
  booktitle    = {{IEEE} Conference on Games (CoG)},
  year         = {2025}
}

@article{tomilin2025mealbenchmarkcontinualmultiagent,
      title={{MEAL: A} Benchmark for Continual Multi-Agent Reinforcement Learning}, 
      author={Tristan Tomilin and Luka van den Boogaard and Samuel Garcin and Bram Grooten and Meng Fang and Yali Du and Mykola Pechenizkiy},
      year={2025},
      journal      = {CoRR},
      volume       = {abs/2506.14990}
}

@article{POET,
  author       = {Rui Wang and
                  Joel Lehman and
                  Jeff Clune and
                  Kenneth O. Stanley},
  title        = {Paired Open-Ended Trailblazer {(POET):} Endlessly Generating Increasingly
                  Complex and Diverse Learning Environments and Their Solutions},
  journal      = {CoRR},
  volume       = {abs/1901.01753},
  year         = {2019},
}

@inproceedings{nethack,
 author = {K\"{u}ttler, Heinrich and Nardelli, Nantas and Miller, Alexander and Raileanu, Roberta and Selvatici, Marco and Grefenstette, Edward and Rockt\"{a}schel, Tim},
 booktitle = {Neural Information Processing Systems (NeurIPS)},
 title = {The {NetHack Learning Environment}},
 year = {2020}
}

@inproceedings{
wolczyk2021continual,
title={Continual {World: A} Robotic Benchmark For Continual Reinforcement Learning},
author={Maciej Wolczyk and Micha{\l} Zaj{\k{a}}c and Razvan Pascanu and {\L}ukasz Kuci{\'n}ski and Piotr Mi{\l}o{\'s}},
booktitle={Neural Information Processing Systems (NeurIPS)},
year={2021},
}

@incollection{LecunnInit,
  author       = {Yann LeCun and
                  L{\'{e}}on Bottou and
                  Genevieve B. Orr and
                  Klaus{-}Robert M{\"{u}}ller},
  title        = {Efficient {BackProp}},
  booktitle    = {Neural Networks: Tricks of the Trade - Second Edition},
  series       = {Lecture Notes in Computer Science},
  volume       = {7700},
  pages        = {9--48},
  publisher    = {Springer},
  year         = {2012}
}

@inproceedings{henderson2019deepreinforcementlearningmatters,
  author       = {Peter Henderson and
                  Riashat Islam and
                  Philip Bachman and
                  Joelle Pineau and
                  Doina Precup and
                  David Meger},
  title        = {Deep Reinforcement Learning That Matters},
  booktitle    = {{AAAI} Conference on Artificial Intelligence (AAAI)},
  year         = {2018}
}

@article{EmpricalDesignforRL,
  author  = {Andrew Patterson and Samuel Neumann and Martha White and Adam White},
  title   = {Empirical Design in Reinforcement Learning},
  journal = {Journal of Machine Learning Research},
  year    = {2024},
  volume  = {25},
  number  = {318},
  pages   = {1--63}
}

@inproceedings{dabney2021temporallyextended,
title={Temporally-Extended {\ensuremath{\varepsilon}}-Greedy Exploration},
author={Will Dabney and Georg Ostrovski and Andre Barreto},
booktitle={International Conference on Learning Representations (ICLR)},
year={2021}
}

@article{khetarpal2022continualreinforcementlearningreview,
  author       = {Khimya Khetarpal and
                  Matthew Riemer and
                  Irina Rish and
                  Doina Precup},
  title        = {Towards Continual Reinforcement Learning: {A} Review and Perspectives},
  journal      = {Journal of Artificial Intelligence Research},
  volume       = {75},
  pages        = {1401--1476},
  year         = {2022}
}

@inproceedings{sokar2023dormantneuronphenomenondeep,
  author       = {Ghada Sokar and
                  Rishabh Agarwal and
                  Pablo Samuel Castro and
                  Utku Evci},
  title        = {The Dormant Neuron Phenomenon in Deep Reinforcement Learning},
  booktitle    = {International Conference on Machine Learning (ICML)},
  year         = {2023}
}

@inproceedings{abbas2023loss,
      title={Loss of plasticity in continual deep reinforcement learning}, 
      author={Zaheer Abbas and Rosie Zhao and Joseph Modayil and Adam White and Marlos C. Machado},
      booktitle={Conference on Lifelong Learning Agents (CoLLAs)},
      year={2023}
}

@inproceedings{powers2022cora,
  author       = {Sam Powers and
                  Eliot Xing and
                  Eric Kolve and
                  Roozbeh Mottaghi and
                  Abhinav Gupta},
  title        = {{CORA:} {B}enchmarks, baselines, and metrics as a platform for continual reinforcement learning agents},
  booktitle    = {Conference on Lifelong Learning Agents (CoLLAs)},
  year         = {2022}
}

@article{bellemare2013arcade,
   title={The {A}rcade {L}earning {E}nvironment: {A}n evaluation platform for general agents},
   volume={47},
   journal={Journal of Artificial Intelligence Research},
   author={Marc G. Bellemare and Yavar Naddaf and Joel Veness and Michael Bowling},
   year={2013}, pages={253–279} }

@article{zakka2025mujocoplayground,
 author       = {Kevin Zakka and
                  Baruch Tabanpour and
                  Qiayuan Liao and
                  Mustafa Haiderbhai and
                  Samuel Holt and
                  Jing Yuan Luo and
                  Arthur Allshire and
                  Erik Frey and
                  Koushil Sreenath and
                  Lueder A. Kahrs and
                  Carmelo Sferrazza and
                  Yuval Tassa and
                  Pieter Abbeel},
  title        = {{MuJoCo} {P}layground},
  journal      = {CoRR},
  volume       = {abs/2502.08844},
  year         = {2025}
}

@article{tassa2018deepmindcontrolsuite,
  author       = {Yuval Tassa and
                  Yotam Doron and
                  Alistair Muldal and
                  Tom Erez and
                  Yazhe Li and
                  Diego de Las Casas and
                  David Budden and
                  Abbas Abdolmaleki and
                  Josh Merel and
                  Andrew Lefrancq and
                  Timothy P. Lillicrap and
                  Martin A. Riedmiller},
  title        = {{DeepMind} Control Suite},
  journal      = {CoRR},
  volume       = {abs/1801.00690},
  year         = {2018}
}

@article{beatie2016deepmindlab,
  author       = {Charles Beattie and
                  Joel Z. Leibo and
                  Denis Teplyashin and
                  Tom Ward and
                  Marcus Wainwright and
                  Heinrich K{\"{u}}ttler and
                  Andrew Lefrancq and
                  Simon Green and
                  V{\'{\i}}ctor Vald{\'{e}}s and
                  Amir Sadik and
                  Julian Schrittwieser and
                  Keith Anderson and
                  Sarah York and
                  Max Cant and
                  Adam Cain and
                  Adrian Bolton and
                  Stephen Gaffney and
                  Helen King and
                  Demis Hassabis and
                  Shane Legg and
                  Stig Petersen},
  title        = {{DeepMind Lab}},
  journal      = {CoRR},
  volume       = {abs/1612.03801},
  year         = {2016}
}

@inproceedings{platanios2020jelly,
  author       = {Emmanouil Antonios Platanios and
                  Abulhair Saparov and
                  Tom M. Mitchell},
  title        = {{J}elly {B}ean {W}orld: {A} testbed for never-ending learning},
  booktitle    = {International Conference on Learning Representations (ICLR)},
  year         = {2020}
}

@inproceedings{zhang2023gobigger,
title={{G}o{B}igger: {A} scalable platform for cooperative-competitive multi-agent interactive simulation},
author={Ming Zhang and Shenghan Zhang and Zhenjie Yang and Lekai Chen and Jinliang Zheng and Chao Yang and Chuming Li and Hang Zhou and Yazhe Niu and Yu Liu},
booktitle={International Conference on Learning Representations (ICLR)},
year={2023}
}

@misc{AgarLE,
  author       = {Jonathan Deaton},
  title        = {{AgarLE: Agar.io OpenAI Gym Learning Environment}},
  year         = {2018},
  note         = {\texttt{https://github.com/jondeaton/AgarLE}. Accessed: 2025-04-09}
}

@inproceedings{abel2023definition,
  author       = {David Abel and
                  Andr{\'{e}} Barreto and
                  Benjamin Van Roy and
                  Doina Precup and
                  Hado Philip van Hasselt and
                  Satinder Singh},
  title        = {A definition of continual reinforcement learning},
  booktitle    = {Neural Information Processing Systems (NeurIPS)},
  year         = {2023}
}

@inproceedings{lewandowski2025world,
  author       = {Alex Lewandowski and Aditya Ramesh and Edan Meyer and Dale Schuurmans and Marlos C. Machado},
  title        = {The world is bigger: A computationally-embedded perspective on the big world hypothesis},
  booktitle    = {Neural Information Processing Systems (NeurIPS)},
  year         = {2025}
}

@article{kumar2023continual,
  author       = {Saurabh Kumar and
                  Henrik Marklund and
                  Ashish Rao and
                  Yifan Zhu and
                  Hong Jun Jeon and
                  Yueyang Liu and
                  Benjamin Van Roy},
  title        = {Continual Learning as Computationally Constrained Reinforcement Learning},
  journal      = {Foundations and Trends in Machine Learning},
  volume       = {18},
  number       = {5},
  pages        = {913--1053},
  year         = {2025}
}

@inproceedings{sutton2007role,
  author       = {Richard S. Sutton and
                  Anna Koop and
                  David Silver},
  title        = {On the role of tracking in stationary environments},
  booktitle    = {International Conference on Machine Learning (ICML)},
  year         = {2007}
}

@article{janjua2024gvfs,
  author       = {Muhammad Kamran Janjua and
                  Haseeb Shah and
                  Martha White and
                  Erfan Miahi and
                  Marlos C. Machado and
                  Adam White},
  title        = {{GVFs} in the real world: Making predictions online for water treatment},
  journal      = {Machine Learning},
  volume       = {113},
  number       = {8},
  pages        = {5151--5181},
  year         = {2024}
}

@inproceedings{todorov2012mujoco,
  author       = {Emanuel Todorov and
                  Tom Erez and
                  Yuval Tassa},
  title        = {{MuJoCo}: {A} physics engine for model-based control},
  booktitle    = {{IEEE/RSJ} International Conference on Intelligent Robots and Systems (IROS)},
  year         = {2012}
}

@article{machado2018revisiting,
  author       = {Marlos C. Machado and
                  Marc G. Bellemare and
                  Erik Talvitie and
                  Joel Veness and
                  Matthew J. Hausknecht and
                  Michael Bowling},
  title        = {Revisiting the {A}rcade {L}earning {E}nvironment: {E}valuation protocols and
                  open problems for general agents},
  journal      = {Journal of Artificial Intelligence Research},
  volume       = {61},
  pages        = {523--562},
  year         = {2018}
}

@inproceedings{tomilin2023coom,
  author       = {Tristan Tomilin and
                  Meng Fang and
                  Yudi Zhang and
                  Mykola Pechenizkiy},
  title        = {{COOM:} {A} game benchmark for continual reinforcement learning},
  booktitle    = {Neural Information Processing Systems (NeurIPS)},
  year         = {2023}
}

@inproceedings{anand2023prediction,
  author       = {Nishanth Anand and
                  Doina Precup},
  title        = {Prediction and control in continual reinforcement learning},
  booktitle    = {Neural Information Processing Systems (NeurIPS)},
  year         = {2023}
}

@article{DBLP:journals/nature/MnihKSRVBGRFOPB15,
  author       = {Volodymyr Mnih and
                  Koray Kavukcuoglu and
                  David Silver and
                  Andrei A. Rusu and
                  Joel Veness and
                  Marc G. Bellemare and
                  Alex Graves and
                  Martin A. Riedmiller and
                  Andreas Fidjeland and
                  Georg Ostrovski and
                  Stig Petersen and
                  Charles Beattie and
                  Amir Sadik and
                  Ioannis Antonoglou and
                  Helen King and
                  Dharshan Kumaran and
                  Daan Wierstra and
                  Shane Legg and
                  Demis Hassabis},
  title        = {Human-level control through deep reinforcement learning},
  journal      = {Nature},
  volume       = {518},
  number       = {7540},
  pages        = {529--533},
  year         = {2015}
}

@inproceedings{SAC,
  author       = {Tuomas Haarnoja and
                  Aurick Zhou and
                  Pieter Abbeel and
                  Sergey Levine},
  title        = {Soft Actor-Critic: Off-Policy Maximum Entropy Deep Reinforcement Learning
                  with a Stochastic Actor},
  booktitle    = {International Conference on Machine Learning (ICML)},
  year         = {2018}
}

@article{PPO,
  author       = {John Schulman and
                  Filip Wolski and
                  Prafulla Dhariwal and
                  Alec Radford and
                  Oleg Klimov},
  title        = {Proximal Policy Optimization Algorithms},
  journal      = {CoRR},
  volume       = {abs/1707.06347},
  year         = {2017}
}

@article{blackwell1962discrete,
author = {David Blackwell},
title = {{Discrete Dynamic Programming}},
volume = {33},
journal = {The Annals of Mathematical Statistics},
number = {2},
pages = {719--726},
year = {1962}
}

@book{puterman2014markov,
  author = {Puterman, Martin L.},
  publisher = {John Wiley \& Sons},
  title = {Markov Decision Processes: Discrete Stochastic Dynamic Programming},
  year = 2014
}

@article{kaelbling1998planning,
  author       = {Leslie Pack Kaelbling and
                  Michael L. Littman and
                  Anthony R. Cassandra},
  title        = {Planning and Acting in Partially Observable Stochastic Domains},
  journal      = {Artificial Intelligence},
  volume       = {101},
  number       = {1-2},
  pages        = {99--134},
  year         = {1998}
}

@inproceedings{javed2024big,
  author          = {Khurram Javed and Richard S. Sutton},
  title           = {The Big World Hypothesis and its Ramifications for Artificial Intelligence},
  year            = 2024,
  booktitle       = {Finding the Frame Workshop at Reinforcement Learning Conference}
}

@inproceedings{anso2019deep,
  author       = {Nil Stolt Ans{\'{o}} and
                  Anton Orell Wiehe and
                  Madalina M. Drugan and
                  Marco A. Wiering},
  title        = {Deep Reinforcement Learning for Pellet Eating in {Agar.io}},
  booktitle    = {International Conference on Agents and Artificial Intelligence (ICAART)},
  year         = {2019}
}

@article{wiehe2018sampled,
  author       = {Anton Orell Wiehe and
                  Nil Stolt Ans{\'{o}} and
                  Madalina M. Drugan and
                  Marco A. Wiering},
  title        = {Sampled Policy Gradient for Learning to Play the Game {Agar.io}},
  journal      = {CoRR},
  volume       = {abs/1809.05763},
  year         = {2018}
}

@article{brockman2016openai,
  author       = {Greg Brockman and
                  Vicki Cheung and
                  Ludwig Pettersson and
                  Jonas Schneider and
                  John Schulman and
                  Jie Tang and
                  Wojciech Zaremba},
  title        = {{OpenAI Gym}},
  journal      = {CoRR},
  volume       = {abs/1606.01540},
  year         = {2016}
}

@article{ba2016layer,
  author       = {Lei Jimmy Ba and
                  Jamie Ryan Kiros and
                  Geoffrey E. Hinton},
  title        = {Layer Normalization},
  journal      = {CoRR},
  volume       = {abs/1607.06450},
  year         = {2016}
}

@article{fujita2021chainer,
  author  = {Yasuhiro Fujita and Prabhat Nagarajan and Toshiki Kataoka and Takahiro Ishikawa},
  title   = {{ChainerRL}: {A} Deep Reinforcement Learning Library},
  journal = {Journal of Machine Learning Research},
  year    = {2021},
  volume  = {22},
  number  = {77},
  pages   = {1-14}
}

@phdthesis{machado2019efficient,
  author = {Marlos C. Machado},
  title = {Efficient Exploration in Reinforcement Learning through Time-Based Representations},
  school = {University of Alberta, Canada},
  year = {2019}
}

@article{dohare2024loss,
  author       = {Shibhansh Dohare and
                  J. Fernando Hernandez{-}Garcia and
                  Qingfeng Lan and
                  Parash Rahman and
                  A. Rupam Mahmood and
                  Richard S. Sutton},
  title        = {Loss of plasticity in deep continual learning},
  journal      = {Nature},
  volume       = {632},
  number       = {8026},
  pages        = {768--774},
  year         = {2024}
}

@article{mesbahi2024tuning,
  author       = {Golnaz Mesbahi and
                  Olya Mastikhina and
                  Parham Mohammad Panahi and
                  Martha White and
                  Adam White},
  title        = {K-percent Evaluation for Lifelong {RL}},
  journal      = {CoRR},
  volume       = {abs/2404.02113},
  year         = {2024}
}

@article{lewandowski2023curvature,
  author       = {Alex Lewandowski and
                  Haruto Tanaka and
                  Dale Schuurmans and
                  Marlos C. Machado},
  title        = {Curvature Explains Loss of Plasticity},
  journal      = {CoRR},
  volume       = {abs/2312.00246},
  year         = {2023}
}

@inproceedings{cho2014properties,
  author       = {Kyunghyun Cho and
                  Bart van Merrienboer and
                  Dzmitry Bahdanau and
                  Yoshua Bengio},
  title        = {On the Properties of Neural Machine Translation: Encoder-Decoder Approaches},
  booktitle    = {EMNLP Workshop on Syntax, Semantics
                  and Structure in Statistical Translation},
  year         = {2014}
}

@inproceedings{ChevalierBoisvert23minigrid,
  author       = {Maxime Chevalier{-}Boisvert and
                  Bolun Dai and
                  Mark Towers and
                  Rodrigo Perez{-}Vicente and
                  Lucas Willems and
                  Salem Lahlou and
                  Suman Pal and
                  Pablo Samuel Castro and
                  Jordan K. Terry},
  title        = {Minigrid {\&} {Miniworld}: {M}odular {\&} Customizable Reinforcement
                  Learning Environments for Goal-Oriented Tasks},
  booktitle    = {Neural Information Processing Systems (NeurIPS)},
  year         = {2023}
}

@inproceedings{johnson2016malmo,
  author       = {Matthew Johnson and
                  Katja Hofmann and
                  Tim Hutton and
                  David Bignell},
  title        = {The {Malmo Platform} for Artificial Intelligence Experimentation},
  booktitle    = {International Joint Conference on Artificial Intelligence (IJCAI)},
  year         = {2016}
}

@inproceedings{Guss2019MineRL,
  author       = {William H. Guss and
                  Brandon Houghton and
                  Nicholay Topin and
                  Phillip Wang and
                  Cayden R. Codel and
                  Manuela Veloso and
                  Ruslan Salakhutdinov},
  title        = {{MineRL}: {A} Large-Scale Dataset of {Minecraft} Demonstrations},
  booktitle    = {International Joint Conference on
                  Artificial Intelligence (IJCAI)},
  year         = {2019}
}

@inproceedings{long2024learning,
  author       = {Junfeng Long and
                  Wenye Yu and
                  Quanyi Li and
                  Zirui Wang and
                  Dahua Lin and
                  Jiangmiao Pang},
  title        = {Learning {H-Infinity} Locomotion Control},
  booktitle    = {Conference on Robot Learning (CoRL)},
  year         = {2024}
}

@InProceedings{Aljundi2018memory,
author = {Aljundi, Rahaf and Babiloni, Francesca and Elhoseiny, Mohamed and Rohrbach, Marcus and Tuytelaars, Tinne},
title = {Memory Aware Synapses: Learning what (not) to forget },
booktitle = {European Conference on Computer Vision (ECCV)},
year = {2018}
}

@article{Kirkpatrick2017overcoming,
author = {James Kirkpatrick  and Razvan Pascanu  and Neil Rabinowitz  and Joel Veness  and Guillaume Desjardins  and Andrei A. Rusu  and Kieran Milan  and John Quan  and Tiago Ramalho  and Agnieszka Grabska-Barwinska  and Demis Hassabis  and Claudia Clopath  and Dharshan Kumaran  and Raia Hadsell },
title = {Overcoming catastrophic forgetting in neural networks},
journal = {Proceedings of the National Academy of Sciences},
volume = {114},
number = {13},
pages = {3521-3526},
year = {2017},
}

@inproceedings{Rolnick2019Experience,
 author = {Rolnick, David and Ahuja, Arun and Schwarz, Jonathan and Lillicrap, Timothy and Wayne, Gregory},
 booktitle = {Neural Information Processing Systems (NeurIPS)},
 title = {Experience Replay for Continual Learning},
 year = {2019}
}

@InProceedings{Li2016Learning,
author="Li, Zhizhong
and Hoiem, Derek",
editor="Leibe, Bastian
and Matas, Jiri
and Sebe, Nicu
and Welling, Max",
title="Learning Without Forgetting",
booktitle="European Conference on Computer Vision (ECCV)",
year="2016"
}

@inproceedings{ash2020warm,
  author       = {Jordan T. Ash and
                  Ryan P. Adams},
  title        = {On Warm-Starting Neural Network Training},
  booktitle    = {Neural Information Processing Systems (NeurIPS)},
  year         = {2020}
}
\bibliographystyle{rlj}

\beginSupplementaryMaterials
\appendix

\section{Broader Implications}
\label{appendix:Broader_Implication}
This work introduces a new evaluation framework for continual RL, built around the video game Agar.io.  Given that this project is based on a video game and designed for benchmarking and evaluation purposes, there are no foreseeable negative societal impacts. As with most research involving simulated environments in AI, the risks are minimal and do not require significant mitigation. \looseness=-1

\section{Use of Large Language Models}

LLMs were only used for minor language polishing during the writing of the paper. 

\section{Software Release and Technical Details}~\label{sec:release}

\par{
    \texttt{AgarCL} is released as free, open-source software under the terms of the MIT license. It is implemented on top of the AgarLE \citep{AgarLE}, an incomplete implementation of Agar.io with an OpenAI Gym~\citep{brockman2016openai} interface. \texttt{AgarCL}'s source code is publicly available in: 
    \begin{center}
    \texttt{https://github.com/machado-research/AgarCL}
    \end{center}
     
    \texttt{AgarCL}'s core simulation engine is implemented in C++, and rendering is handled by OpenGL. A Python interface, built with Pybind11, is also supported. In our experiments, when looking at the interquantile mean (IQM) over ten independent trials, a random agent, written in Python, receives {$2{,}016$} frames per second with a frame skip of $1$. With a frame skip of $4$ (default), the agent receives {$1{,}163$} observations per second, which represents {$4{,}652$} frames in the game and {$1{,}163$} actions selected by the agent. We report the IQM to discard fluctuations due to other jobs in the system. This experiment used $2$ cores of an \textit{Intel Xeon Gold 6448Y} CPU and one \textit{NVIDIA L40s} GPU. Correspondingly, the IQM when evaluating \texttt{AgarCL} without a GPU was $1{,}440$ and $1{,}092$, respectively. \looseness=-1

}
\par{

}

\section{Implementation Details in AgarCL}
\label{Env_details}

In this section, we discuss in more detail the full-game experiment and the implementation of the fixed-policy bots.
\subsection{\texttt{AgarCL} 
Experiment Details}
\label{appendix:full-game_experiment_detail}

In our main evaluation, we use a $350\times350$ arena containing $500$ pellets, $10$ viruses, and $8$ heuristic bots. Upon being consumed, any agent or bot respawns immediately with an initial mass of $25$, while all other entities preserve their current state. We describe the bot heuristics in Appendix~\ref{appendix:bot_description}.

Every $600$ ticks ($150$ ticks if we use frame skipping of 4), pellets and viruses that have been consumed or destroyed are regenerated with a uniform probability across the arena. This maintains fixed totals of $500$ and $10$, respectively. Mass decay is applied every $60$ ticks. We conduct each trial for $160\times10^{6}$ frames. 

\subsection{Fixed-Policy Bots}
\label{appendix:bot_description}
We implemented four heuristic bots in \texttt{AgarCL}: Aggressive, Aggressive‑Shy, Hungry, and Hungry‑Shy. We did so to explore how simple rules shape emergent game dynamics. All bots follow a fixed policy that always targets the nearest pellet, but differ in how they interact with other agents/bots. This set of behaviours allows us to assess how different action priors influence both individual success and the overall balance of the game (see Appendix~\ref{appendix:other_agents}). The four bots' policies are: \looseness=-1

\begin{itemize}
    \item Aggressive: This bot first looks for any smaller opponent within a defined radius and attempts to consume it; if no suitable target is found, it switches to pellet collection. \looseness-1
    \item Aggressive‑Shy: Like the Aggressive bot, it will hunt smaller opponents, but if a larger opponent approaches within its ``shy'' radius, it immediately flees and only returns to hunting once the threat has passed. \looseness-1

    \item Hungry: This bot ignores other players entirely and chases the closest pellet at every step.

    \item Hungry‑Shy: Focused on pellet foraging like the Hungry bot, it additionally monitors for larger opponents: if one comes too close, it retreats before resuming its hunt.

\end{itemize}

\section{Additional Related Work}
\label{sec_RelatedWork}




\paragraph{Continual Reinforcement Learning.}

\par{
There have been many attempts to formalize the continual reinforcement learning (CRL) problem, each highlighting different aspects of non-stationarity and lifelong adaptation. \citet{khetarpal2022continualreinforcementlearningreview} introduce a taxonomy of continual reinforcement learning (CRL) that focuses on two fundamental dimensions of non-stationarity: `scope', referring to the extent of variation across tasks or domains, and `driver', representing the main source of change. Building on this, \citet{abel2023definition} define CRL as an ongoing adaptation process, in which agents continuously update their policies in response to evolving objectives, dynamics, or reward structures. More recently, \citet{kumar2023continual} define continual learning under computational constraints by proposing a framework that maintains a balance between past knowledge and the efficiency of online updates over an extended period. Collectively, these CRL formalisms establish foundational principles that guide the design of benchmarks and algorithms. 
}

\paragraph{Existing Environments.}
\par{Traditional RL testbeds \cite[e.g.,][]{zakka2025mujocoplayground, tassa2018deepmindcontrolsuite} have driven rapid progress in RL, but they are not tailored for CRL. Accordingly, many CRL studies resort to switching among different games to induce non-stationarity \cite[e.g.,][]{abbas2023loss}, relying on clearly defined train-test boundaries and assuming a well-structured notion of tasks and episodes. As we argue in this paper, slower, smoother changes capture a different (and underexplored) type of non-stationarity in the problems that motivated CRL. JellyBeanWorld~\citep{platanios2020jelly} is the main platform for CRL in non-episodic settings that we are aware of. In JellyBeanWorld, agents navigate an infinite two-dimensional grid, interacting with various items by collecting or avoiding them. The agent's states are partially observable. Although JellyBeanWorld is clearly valuable for CRL research, it offers simpler observations and (discrete) dynamics, without the ever-changing nature of AgarCL.}

\paragraph{Agar-Like Environments.} \par{Most of the results relying on Agar-like environments for evaluation consisted of assessing basic agent capabilities, mostly related to pellet eating~\citep{wiehe2018sampled,anso2019deep}, without ever putting forward the environment as a key artifact. To the best of our knowledge, GoBigger~\citep{zhang2023gobigger} is the only other actual evaluation framework for AI research based on the Agar.io game. It shares certain surface-level similarities with our own, particularly in its Agar-style gameplay mechanics, but it was introduced to support a fundamentally different problem. Its main goal is to provide a platform to study collective behaviours in multi-agent reinforcement learning. Thus, its features include supporting multiple teams of agents (in the original game, collaboration should emerge from communication) in settings where agents are organized into a few teams. The larger environment configuration supports $24$ agents organized into four teams ($6$ each). The game is designed to be \emph{episodic}, with larger maps artificially divided into episodes that have at most $14,400$ frames ($12$ minutes). These are all in direct contrast to the continual learning problem we focus on, with potentially unbounded episodes and no pre-defined teams, but with a much bigger number of agents. GoBigger also does not support pixel-based observations; it just supports something akin to our symbolic observations with information about objects' positions and velocities. Finally, we benchmarked the frames per second (FPS) over ten independent trials using the \texttt{st\_t6p4}\footnote{This configuration typically includes 1000 pellets, a 144$\times$144 arena, and 14,400 frames per episode. We modified the GoBigger implementation to support a single player.} configuration---the largest map setting in the GoBigger implementation. Our environment significantly outperforms GoBigger in simulation speed, achieving an interquartile mean (IQM) of $4,212$ fps with GoBigger-style observations, compared to GoBigger’s IQM of $205$ fps under the same observation setup. This experiment was run using $2$ cores of an \textit{Intel Xeon Gold 6448Y} CPU.} \looseness=-1


\section{Algorithm Details}~\label{appendix:algorithms}

In this section, we focus on the adaptations we have made to DQN, PPO, and SAC to allow them to work somewhat effectively in \texttt{AgarCL}.

\paragraph{DQN.} We used PFRL's DQN implementation \citep{fujita2021chainer}. We did so by discretizing the environment's continuous actions. We ended up with 24 actions: 8 directions times the 3 discrete actions. The predefined directions were: \textsc{up} \((0,1)\), \textsc{up-right} \((1,1)\), \textsc{right} \((1,0)\), \textsc{down-right} \((1,-1)\), \textsc{down} \((0,-1)\), \textsc{down-left} \((-1,-1)\), \textsc{left} \((-1,0)\), and \textsc{up-left} \((-1,1)\). \looseness=-1

As discussed throughout, the resets inherent to episodic tasks can benefit exploration by allowing agents to recover from ``bad'' states. However, in the face of continual problems, the agent must naturally recover from a ``bad'' state. In our experiments, we noticed that the $\epsilon$-greedy strategy was ineffective in the non-episodic tasks we considered. Thus, in such tasks, we introduced temporally-extended exploration~\citep{machado2019efficient} via $\epsilon$z-greedy~\citep{dabney2021temporallyextended}, which achieves temporal persistence by extending random actions over multiple steps using a heavy-tailed duration distribution. \looseness=-1

\paragraph{PPO.} Our implementation extends PFRL's to the hybrid action setting. We adopted a shared neural network with two heads: an \textit{actor} head that outputs action probabilities and a value \textit{critic} head that estimates state values. Preliminary results showed that this architecture was more effective than two independent networks. The actor head splits a $256$-dimensional feature vector into discrete and continuous branches: a softmax head for categorical action probabilities and a Gaussian head (state-independent covariance) for continuous action means and variances, sampling both at each timestep to form a factored joint policy. The critic head is a compact one‐layer MLP producing a scalar \(V(s)\), trained with PPO’s clipped value‐function loss and generalized advantage estimation. Finally, we relied on PPO's entropy regularization for exploration.

\paragraph{SAC.} Similarly to PPO, we extended PFRL's SAC implementation to the hybrid action setting and relied on the algorithm's entropy-maximizing term for exploration. Unlike PPO’s single-value head, our implementation comprises three fully independent networks—one actor and two critics—each with its own encoder, $\phi$, that directly processes the raw observation. The actor mirrors the PPO policy architecture, but with an entropy-regularized objective. Likewise, each critic encodes the observation using $\phi$. The resulting representation is then concatenated with the continuous action, after which the critic predicts Q-values for each discrete action. The Q-values are selected with respect to the actor’s sampled discrete action.
By structuring the critics this way, we avoid the combinatorial explosion that would arise if we naively input every possible hybrid action pair, and we prevent gradient interference between discrete and continuous parameters.

\section{Tuning Details}
\label{appendix:training-details}
Hyperparameter tuning is especially critical in reinforcement learning due to the inherent instability and sensitivity of RL algorithms. Unlike supervised learning, RL involves exploration-exploitation trade-offs, nonstationary data distributions, and delayed rewards, all of which can magnify the effects of poorly chosen hyperparameters. Parameters such as step size, discount factor, exploration noise, entropy regularization (in policy gradient methods), and update frequency can significantly influence learning dynamics and final policy performance. Improper tuning can lead to divergence, suboptimal policies, or excessive performance variance.
\subsection{DQN Tuning}
\label{appendix:dqn_tuning}
For tuning DQN, we swept over the hyperparameters listed in Table \ref{tab:dqn_hyperparams}. As discussed in the paper, both the agent's network and target network consist of three convolutional layers, each followed by a ReLU activation and a Layer Normalization layer. In preliminary experiments, we found that omitting Layer Normalization prevented the agent from learning altogether. All weights were initialized using a LeCun normal initialization \citep{LecunnInit}. Each hyperparameter combination was trained using three different random seeds. After identifying the best hyperparameter configuration, we conducted ten additional independent runs using the best‐performing configuration (see Table~\ref{tab:dqn_best_hypers} on the next page). As shown in Table~\ref{tab:dqn_hyperparams}, also on the next page, the discount factor, $\gamma$, the number of epochs, the replay buffer size, and the target‐network update interval were each fixed at a single value, while the remaining hyperparameters were swept. Finally, we applied the hyperparameter settings from continual \textsc{mini-game~4} to train the full‐game agent.

\begin{table}[th]
\centering
\caption{Values of hyperparameters that we swept over when tuning DQN.}
\label{tab:dqn_hyperparams}

\begin{tabular}{ll}
\hline
\textbf{Hyperparameter} & \textbf{Values / Settings} \\
\hline
Step size (\texttt{--lr}) & $10^{-5}$, $3 \cdot10^{-5}$, $10^{-4}$, $3\cdot10^{-4}$ \\
Batch Accumulator (\texttt{--batch\_accumulator}) & \texttt{"sum"}, \texttt{"mean"} \\
Soft Update Coefficient (\texttt{--tau}) & $10^{-2}$, $5^{-3}$ \\
Batch Size (\texttt{--batch-size}) & \texttt{32}, \texttt{64} \\
Replay Buffer Size (\texttt{--replay-buffer}) & $10^{5}$ \\
Number of Epochs (\texttt{--epochs}) & \texttt{1} \\
Target Network Update Interval (\texttt{--update\_interval}) & \texttt{4} \\
Gamma (\texttt{--$\gamma$}) & \texttt{0.99} \\
Exploration algorithm &\texttt{$\epsilon$-Greedy}, \texttt{$\epsilon$z-Greedy}\\
\hline
\end{tabular}

\end{table}

\begin{table}[th]
  \centering
  \caption{Best hyperparameters for DQN on each minigame.} 
  \label{tab:dqn_best_hypers}

  {\footnotesize
  \resizebox{1.05\textwidth}{!}{%
    \begin{threeparttable}
      \renewcommand{\arraystretch}{1.2}
      \begin{tabular}{llccccc}
        \toprule
        \multirow{2}{*}{\textbf{Category}} 
          & \multirow{2}{*}{\textbf{Mini-game}} 
          & \multicolumn{4}{c}{\textbf{Hyperparameters}} 
          & \multirow{2}{*}{\textbf{Exploration Algorithm}} \\
        \cmidrule(l){3-6}
          & 
          & \texttt{Step Size} 
          & \texttt{Batch Accumulator} 
          & \texttt{Soft Update Coefficient} 
          & \texttt{Batch Size} 
          & \\
         \midrule
        \multirow{6}{*}{Episodic}
          & \textcircled{1} & $10^{-4}$       & mean & $5\cdot10^{-5}$ & $64$ & $\epsilon$-Greedy \\
          & \textcircled{2} & $3\cdot10^{-4}$ & sum  & $5\cdot10^{-5}$ & $32$ & $\epsilon$-Greedy \\
          & \textcircled{3} & $10^{-4}$       & mean & $5\cdot10^{-5}$ & $32$ & $\epsilon$-Greedy \\
          & \textcircled{4} & $3\cdot10^{-4}$ & sum  & $10^{-2}$       & $32$ & $\epsilon$-Greedy \\
          & \textcircled{5} & $10^{-5}$       & mean & $10^{-2}$       & $64$ & $\epsilon$-Greedy \\
          & \textcircled{6} & $10^{-5}$       & mean & $10^{-2}$       & $64$ & $\epsilon$-Greedy \\
        \addlinespace
        \multirow{6}{*}{Continuing}
          & \textcircled{1} & $10^{-4}$       & sum  & $5\cdot10^{-5}$ & $32$ & $\epsilon$z-Greedy \\
          & \textcircled{2} & $3\cdot10^{-4}$ & sum  & $10^{-2}$       & $32$ & $\epsilon$z-Greedy \\
          & \textcircled{3} & $10^{-4}$       & mean & $5\cdot10^{-5}$ & $32$ & $\epsilon$z-Greedy \\
          & \textcircled{4} & $10^{-4}$       & sum  & $5\cdot10^{-5}$ & $32$ & $\epsilon$z-Greedy \\
          & \textcircled{5} & $10^{-4}$       & mean & $10^{-2}$       & $32$ & $\epsilon$z-Greedy \\
          & \textcircled{6} & $3\cdot10^{-4}$ & mean & $10^{-2}$       & $32$ & $\epsilon$z-Greedy \\
        \midrule
        \multirow{4}{*}{Other Agents}
          & \textcircled{7}\texttt{-Small-Sparse}\tnote{1} & $10^{-5}$       & mean & $5\times10^{-2}$       & $32$ & $\epsilon$-Greedy \\
          & \textcircled{7}\texttt{-Normal-Dense}\tnote{2} & $3\times10^{-4}$ & mean & $10^{-2}$ & $64$ & $\epsilon$-Greedy \\
           & \textcircled{8}\texttt{-Small-Sparse}\tnote{1} & $10^{-5}$       & mean & $5\times10^{-2}$       & $64$ & $\epsilon$-Greedy \\
          & \textcircled{8}\texttt{-Normal-Dense}\tnote{2} & $1\times10^{-5}$ & sum & $10^{-2}$ & $64$ & $\epsilon$-Greedy \\
          \midrule
          \multirow{1}{*}{Virus}
          & \textcircled{9} & $3\times10^{-4}$& sum & $10^{-2}$       & $128$ & $\epsilon$-Greedy \\
        \bottomrule
      \end{tabular}
      \begin{tablenotes}
        \footnotesize
        \item[1] Limited (Sparse) arena: $200\times200$ with 200 randomly respawned pellets.
        \item[2] Normal (Dense) arena:  $350\times350$ with 500 randomly respawned pellets.
      \end{tablenotes}
    \end{threeparttable}
  }}
\end{table}

\subsection{PPO Tuning}
Hyperparameter tuning for PPO is particularly challenging due to the size of the configuration space: minimally covering relevant combinations requires at least $324$ runs per experiment. We employed a shared neural network to learn both actor and critic, using the same architecture described in Appendix~\ref{appendix:algorithms}. All network weights are initialized using LeCun normal initialization \citep{LecunnInit}. The policy head is a Gaussian distribution with a state-independent, learned covariance—a configuration that is standard in PPO implementations \cite[e.g.,][]{henderson2019deepreinforcementlearningmatters}. 

In Table~\ref{tab:ppo_hyperparams}, we have the choice between two reward normalization schemes as a hyperparameter. The first is \emph{min–max normalization}, which linearly rescales the raw reward \(r_t\) into the range \([-1, 1]\):
\begin{equation}
\tilde r_t^{\text{(min–max)}} = \frac{r_t - r_{\min}}{\,r_{\max} - r_{\min} + \epsilon\,},
\label{eq:minmax}
\end{equation}

where \(\epsilon > 0\) is a small constant added for numerical stability.

The second method, \emph{variance normalization}, uses an exp. weighted moving average of the returns: \looseness=-1
\noindent
\begin{minipage}{0.48\textwidth}
\begin{equation}
G_t = \gamma\,G_{t-1} + r_t,
\label{eq:return_ema}
\end{equation}
\end{minipage}\hfill
\begin{minipage}{0.48\textwidth}
\begin{equation}
\tilde r_t^{\text{(var--norm)}} = \frac{r_t}{\sqrt{\mathrm{Var}[G_t] + \epsilon}},
\label{eq:varnorm}
\end{equation}
\end{minipage}

where \(G_t\) is the smoothed return, and normalization is performed by dividing the reward by the square root of its running variance. This ensures that the normalized reward maintains approximately unit variance under an exponential moving average with discount factor \(\gamma\). After normalization, the reward is clipped to lie within the fixed range \([-10, 10]\) to limit the effect of outliers during training:
\begin{equation}
\tilde r_t^{\text{(clipped)}} = \mathrm{clip}\left(\tilde r_t^{\text{(var--norm)}}, -10, 10\right).
\end{equation}

Initially, we swept over different hyperparameter combinations for PPO in some episodic settings following Table ~\ref{tab:ppo_hyperparams}. Through these experiments, we observed that a value function coefficient of $0.9$, an update interval of $5000$, step sizes of either $10^{-5}$ or $3 \times 10^{-5}$, and epochs set to either $10$ or $15$ yielded consistently strong performance. Based on sensitivity analyses across most mini-game tasks, we narrowed down the range of hyperparameters we swept over for the other tasks. The smaller set of hyperparameters we swept over is shown in Table~\ref{tab:ppo_hyperparams_2}. Table ~\ref{tab:ppo_best_hypers} shows the best-performing hyperparameters that we used in each task at the end. Finally, we applied the hyperparameter settings from continual \textsc{mini-game~4} to train the full‐game agent.

\begin{table}[H]
\centering
\caption{Values of hyperparameters that we swept over when tuning PPO.}

\label{tab:ppo_hyperparams}
\begin{tabular}{ll}
\toprule
\textbf{Hyperparameter} & \textbf{Values / Settings} \\
\midrule
Reward function (\texttt{--reward})        
  & \texttt{min\_max}, \texttt{variance\_norm}\\
Step size (\texttt{--lr})              
  & $10^{-5}$, $3\times10^{-5}$, $3\times10^{-4}$, $10^{-4}$ \\
Epochs (\texttt{--epochs})                  
  & 10, 15, 20 \\
Max gradient norm (\texttt{--max-grad-norm}) 
  & 0.5, 0.7, 0.9 \\
Entropy coefficient (\texttt{--entropy-coef}) 
  & 0.05, 0.01, 0.1, 0.5 \\
Clipping epsilon (\texttt{--clip-eps})       
  & 0.2, 0.4 \\
Discount factor (\texttt{--gamma})           
  & 0.995 \\
GAE parameter (\texttt{--lambda})            
  & 0.97 \\
Value‐function coefficient (\texttt{--value-func-coef}) 
  & 0.9, 0.5 \\
Batch size (\texttt{--batch-size})           
  & 64 \\
Update interval (\texttt{--update-interval})  
  & $1024$, $2048$, $5000$ \\
\bottomrule
\end{tabular}

\end{table}

\begin{table}[H]
\centering
\caption{Updated PPO hyperparameters used for tuning.}
\label{tab:ppo_hyperparams_2}
\begin{tabular}{ll}
\toprule
\textbf{Hyperparameter} & \textbf{Values / Settings} \\
\midrule
Reward function (\texttt{--reward})        
  & \texttt{min\_max}, \texttt{variance\_norm}\\
Step size (\texttt{--lr})              
  & $10^{-5}$, $3\times10^{-5}$ \\
Epochs (\texttt{--epochs})                  
  & 10, 15 \\
Max gradient norm (\texttt{--max-grad-norm}) 
  & 0.5, 0.7, 0.9 \\
Entropy coefficient (\texttt{--entropy-coef}) 
  & 0.05, 0.01, 0.1, 0.5 \\
Clipping epsilon (\texttt{--clip-eps})       
  & 0.2, 0.4 \\
Discount factor (\texttt{--gamma})           
  & 0.995 \\
GAE parameter (\texttt{--lambda})            
  & 0.97 \\
Value‐function coefficient (\texttt{--value-func-coef}) 
  & 0.9 \\
Batch size (\texttt{--batch-size})           
  & 64 \\
Update interval (\texttt{--update-interval})  
  & 5000 \\
\bottomrule
\end{tabular}
\end{table}

\begin{table}[H]
  \centering
  \caption{Best hyperparameters for PPO on each minigame.}
  \label{tab:ppo_best_hypers}
  \resizebox{1.05\textwidth}{!}{%
    {\footnotesize
    \begin{threeparttable}
      \renewcommand{\arraystretch}{1.2}
      \begin{tabular}{llcccccc}
        \toprule
        \multirow{2}{*}{\textbf{Category}} 
          & \multirow{2}{*}{\textbf{Mini-game}} 
          & \multicolumn{6}{c}{\textbf{Hyperparameters}} \\
        \cmidrule(l){3-8}
          & 
          & \texttt{Reward Function} 
          & \texttt{Step Size} 
          & \texttt{Epochs} 
          & \texttt{Max Grad Norm} 
          & \texttt{Entropy Coef} 
          & \texttt{Clip Eps} \\
        \midrule
        \multirow{6}{*}{Episodic}
          & \textcircled{1} & \texttt{Min Max}      & $10^{-4}$        & 15 & 0.7  & 0.01 & 0.4 \\
          & \textcircled{2} & \texttt{Min Max}      & $10^{-5}$        & 10 & 0.7  & 0.01 & 0.2 \\
          & \textcircled{3} & \texttt{Min Max}      & $10^{-5}$        & 10 & 0.9  & 0.01 & 0.2 \\
          & \textcircled{4} & \texttt{Min Max}      & $10^{-5}$        & 15 & 0.9  & 0.05 & 0.4 \\
          & \textcircled{5} & \texttt{Min Max}      & $10^{-4}$        & 10 & 0.7  & 0.01 & 0.4 \\
          & \textcircled{6} & \texttt{Min Max}      & $3\times10^{-5}$ & 10 & 0.5  & 0.05 & 0.2 \\
        \addlinespace
        \multirow{6}{*}{Continuing}
          & \textcircled{1} & \texttt{Min Max} & $10^{-4}$       & 4  & 0.5  & 0.005 & 0.2 \\
          & \textcircled{2} & \texttt{Variance Norm} & $10^{-5}$       & 10 & 0.5  & 0.05  & 0.4 \\
          & \textcircled{3} & \texttt{Min Max} & $10^{-4}$       & 4  & 0.5  & 0.01  & 0.3 \\
          & \textcircled{4} & \texttt{Variance Norm} & $10^{-5}$       & 10 & 0.5  & 0.1   & 0.4 \\
          & \textcircled{5} & \texttt{Min Max}       & $3\times10^{-5}$& 10 & 0.7  & 0.05  & 0.4 \\
          & \textcircled{6} & \texttt{Min Max}       & $3\times10^{-4}$& 15 & 0.7  & 0.05  & 0.2 \\
        \midrule
        \multirow{4}{*}{Other Agents}
          & \textcircled{7}-Small-Sparse\tnote{1}  & \texttt{Min Max}       & $3\times10^{-5}$ & 10 & 0.5 & 0.05 & 0.4 \\
          & \textcircled{7}-Normal-Dense\tnote{2}  & \texttt{Variance Norm} & $10^{-5}$        & 10 & 0.9 & 0.1  & 0.4 \\
          & \textcircled{8}-Small-Sparse\tnote{1}  & \texttt{Min Max}       & $3\times10^{-5}$ & 10 & 0.5 & 0.05 & 0.2 \\
          & \textcircled{8}-Normal-Dense\tnote{2}  & \texttt{Min-Max} & $3\times10^{-5}$        & 10 & 0.7 & 0.05  & 0.2 \\
        \midrule
        \multirow{1}{*}{Virus}
          & \textcircled{9} & \texttt{Min Max}       & $10^{-5}$ & 10 & 0.7 & 0.01 & 0.4 \\
        \bottomrule
      \end{tabular}
      \begin{tablenotes}
        \footnotesize
        \item[1] Uses a $200\times200$ arena with 200 randomly respawned pellets.
        \item[2] Uses a $350\times350$ arena with 500 randomly respawned pellets.
      \end{tablenotes}
    \end{threeparttable}
    } 
  }
\end{table}

\subsection{SAC Tuning}
We swept over the hyperparameter combinations in Table~\ref{tab:sac_hyperparams}. In SAC, we employed three separate networks. All network weights were initialized using LeCun normal initialization \citep{LecunnInit}. Table~\ref{tab:sac_best_hypers} summarizes the best-performing hyperparameters across our tasks. Also, we applied the hyperparameter settings from continual \textsc{mini-game~4} to train the full‐game agent.

\begin{table}[H]
\centering
\caption{Values of hyperparameters that we swept over when tuning SAC.}
\label{tab:sac_hyperparams}

\begin{tabular}{ll}
\toprule
\textbf{Hyperparameter} & \textbf{Values / Settings} \\
\midrule
Step size (\texttt{--lr})                   & $10^{-4}$, $3\times10^{-5}$,$10^{-5}$ \\
Reward function (\texttt{--reward})             & \texttt{min\_max}, \texttt{variance\_norm} \\
Replay buffer size (\texttt{--replay-buffer})   & $10^5$ \\
Soft update coefficient (\texttt{--tau})        & $10^{-2}$, $5\times10^{-3}$, $10^{-3}$ \\
Max gradient norm (\texttt{--max-grad-norm})    & $0.5$, $0.7$, $0.9$ \\
Temperature step size (\texttt{--temperature-lr}) & $10^{-4}$, $3\times10^{-4}$ \\
Update interval (\texttt{--update-interval})    & $4$ \\
Batch size (\texttt{--batch-size})              & $64$ \\
Discount factor (\texttt{-\-gamma})              & $0.99$ \\
\bottomrule
\end{tabular}
\end{table}

\begin{table}[H]
  \centering
  \caption{Best hyperparameters for SAC on each minigame.}
  \label{tab:sac_best_hypers}

  \resizebox{1.05\textwidth}{!}{%
    {\footnotesize   
    \begin{threeparttable}
      \renewcommand{\arraystretch}{1.2}
      \begin{tabular}{llccccc}
        \toprule
        \multirow{2}{*}{\textbf{Category}} 
          & \multirow{2}{*}{\textbf{Mini-game}} 
          & \multicolumn{5}{c}{\textbf{Hyperparameters}} \\
        \cmidrule(l){3-7}
          & 
          & \texttt{Reward Function} 
          & \texttt{Step Size} 
          & \texttt{Soft Update Coefficient} 
          & \texttt{Temperature LR} 
          & \texttt{Max Grad Norm} \\
        \midrule
        \multirow{6}{*}{Episodic}
          & \textcircled{1} & \texttt{Min Max}       & $3\times10^{-5}$  & $0.001$ & $10^{-4}$ & $0.7$ \\
          & \textcircled{2} & \texttt{Min Max}       & $3\times10^{-5}$  & $0.001$ & $10^{-4}$ & $0.7$ \\
          & \textcircled{3} & \texttt{Min Max}       & $10^{-4}$         & $0.01$  & $10^{-4}$ & $0.7$ \\
          & \textcircled{4} & \texttt{Min Max}       & $10^{-5}$         & $0.001$ & $10^{-4}$ & $0.7$ \\
          & \textcircled{5} & \texttt{Min Max}       & $10^{-5}$         & $0.001$ & $10^{-4}$ & $0.7$ \\
          & \textcircled{6} & \texttt{Min Max}       & $10^{-4}$         & $0.001$ & $10^{-4}$ & $0.5$ \\
        \addlinespace
        \multirow{6}{*}{Continuing}
          & \textcircled{1} & \texttt{Variance Norm} & $3\times10^{-5}$  & $0.001$ & $10^{-4}$ & $0.5$ \\
          & \textcircled{2} & \texttt{Variance Norm} & $3\times10^{-5}$  & $0.001$ & $10^{-4}$ & $0.7$ \\
          & \textcircled{3} & \texttt{Variance Norm} & $3\times10^{-5}$  & $0.005$ & $10^{-4}$ & $0.9$ \\
          & \textcircled{4} & \texttt{Variance Norm} & $10^{-5}$         & $0.01$  & $10^{-4}$ & $0.5$ \\
          & \textcircled{5} & \texttt{Variance Norm} & $10^{-5}$         & $0.001$ & $10^{-4}$ & $0.7$ \\
          & \textcircled{6} & \texttt{Variance Norm} & $3\times10^{-5}$  & $0.005$ & $10^{-5}$ & $0.9$ \\
        \midrule
        \multirow{4}{*}{Other Agents}
          & \textcircled{7}-Small-Sparse\tnote{1}  & \texttt{Variance Norm} & $10^{-4}$ & $0.01$ & $10^{-4}$ & $0.9$ \\
          & \textcircled{7}-Normal-Dense\tnote{2}  & \texttt{Variance Norm} & $10^{-4}$ & $0.01$ & $10^{-4}$ & $0.7$ \\
          & \textcircled{8}-Small-Sparse\tnote{1}  & \texttt{Variance Norm} & $10^{-5}$ & $0.001$ & $10^{-4}$ & $0.9$ \\
          & \textcircled{8}-Normal-Dense\tnote{2}  & \texttt{Variance Norm} & $10^{-5}$ & $0.005$ & $10^{-4}$ & $0.9$ \\
        \midrule
           \multirow{1}{*}{Virus}
          & \textcircled{9} & \texttt{Min Max} & $10^{-5}$ & $0.01$ & $10^{-4}$ & $0.7$ \\
        \bottomrule
      \end{tabular}
      \begin{tablenotes}
        \footnotesize
        \item[1] Sparse setting: $200\times200$ arena with 200 pellets.
        \item[2] Dense setting: $350\times350$ arena with 500 pellets.
      \end{tablenotes}
    \end{threeparttable}
    } 
  }
\end{table}

\section{Mini-Game Results}

In this section, we provide visualizations of the mini-games used in our evaluation (Section~\ref{sec:mini-games}) and the complete set of results mentioned in the main paper. Those include results in the continual pellet collection tasks (Section~\ref{sec:continual_minigames}), including analyses over the algorithms' hyperparameter sensitivity (Section~\ref{appendix:hyper_sensitivity}) and results in those mini-games when augmenting PPO with a GRU (Section~\ref{sec:experiments_rnns}). Additionally, we report results from mini-games in which the agent faced another bot (Section~\ref{appendix:other_agents}) and those in which it was expected to successfully interact with viruses (Section~\ref{sec:viruses_minigame}).

\subsection{Illustrative Figures of the Pellet-Collection Mini-Games}~\label{sec:mini-games}

\begin{figure}[h]
\begin{center}
  \includegraphics[width=\textwidth]{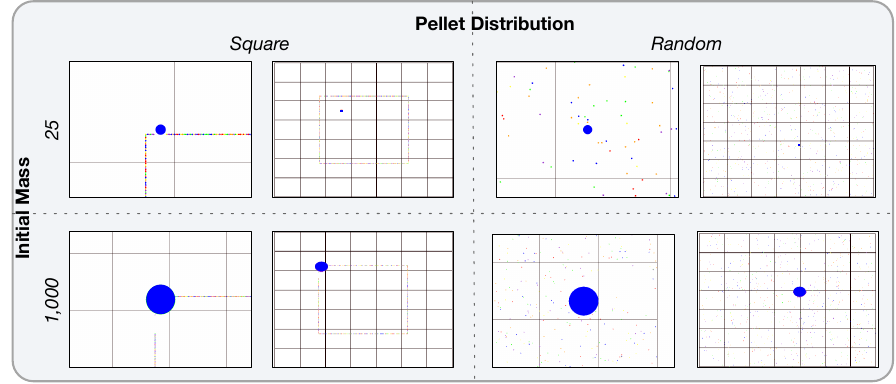}
\caption{Pellet Collection mini games. They were defined in terms of the pellet distribution in the arena, the agent's initial mass, and whether the agent's mass decays (which is hard to depict in an image). Note that we do not have eight mini-games because starting with a mass of 1,000 and no mass decay is an uninteresting setting. In each quadrant, the agent's initial view in the mini-game is shown on the left, and, to provide a sense of scale, a zoomed-out perspective of the same setting is shown on the right. The agent is depicted in blue. \looseness=-1 
}
  \label{fig:agar_mini_games}
  \end{center}
\end{figure}

The mini-games for pellet collection are divided into two sets: Square and Random. The Square mini-games require the agent to collect pellets along a square-shaped path, and it has three versions. The simplest version involves collecting pellets only, with no additional challenges. The second introduces mass decay, and in the third, the agent starts with a much bigger mass (1000 instead of 25). The set of mini-games with randomly scattered pellets uses the same variations. New pellets appear every 600 environment ticks. Figure~\ref{fig:agar_mini_games} depicts all these variants.

\subsection{Continual Problems}~\label{sec:continual_minigames}

We discussed this setting in detail in the main paper; however, due to space constraints, we were unable to present the complete set of results there. They are available in Figure~\ref{fig:uniform_pellets_cont} below.

\begin{figure}[h]
\begin{center}
\includegraphics[width=\textwidth]{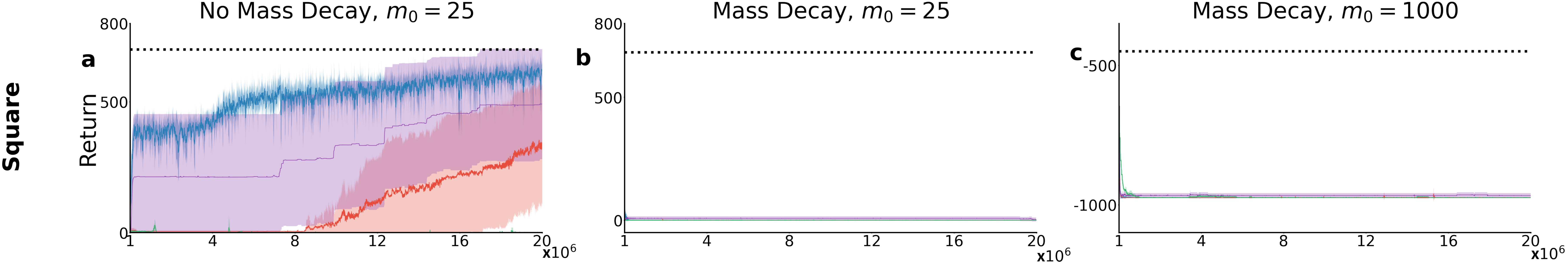}
\includegraphics[width=\textwidth]{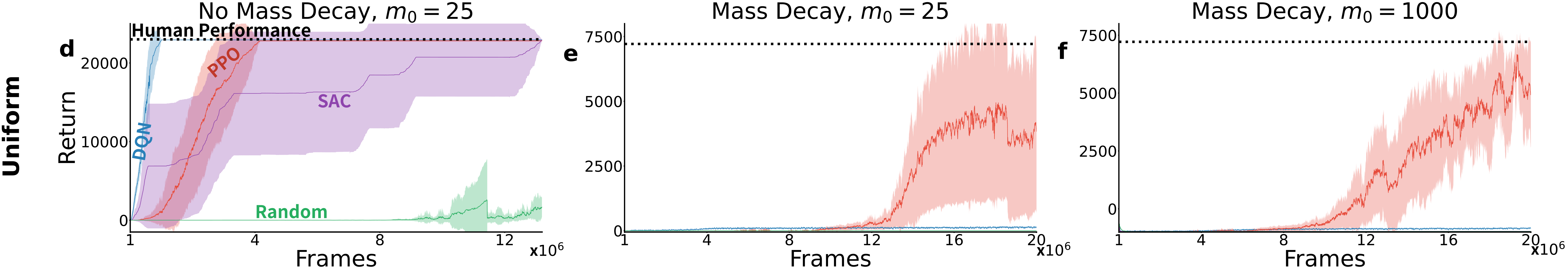}
\caption{Performance of RL methods on \emph{continual} pellet-collection mini-games. Panels \textcircled{a}, \textcircled{b}, and \textcircled{c} show the performance on \emph{the square-path tasks} (mini-games $1$, $2$, and $3$), while \textcircled{d}~,\textcircled{e}, and \textcircled{f} show the performance on \emph{randomly regenerated tasks} (mini-games $4$, $5$, and $6$). Note that the y-axis scales vary across plots. The dashed line marks human performance, and the green line marks the random policy. The shaded region shows the 95\% CI over 10 runs, computed using the t-distribution. \looseness=-1}
\label{fig:uniform_pellets_cont}
\end{center}
\end{figure}

\subsubsection{Hyperparameter Sensitivity Across Mini-games}~\label{appendix:hyper_sensitivity}

\begin{figure}[ht]
\begin{center}
\includegraphics[width=\textwidth]{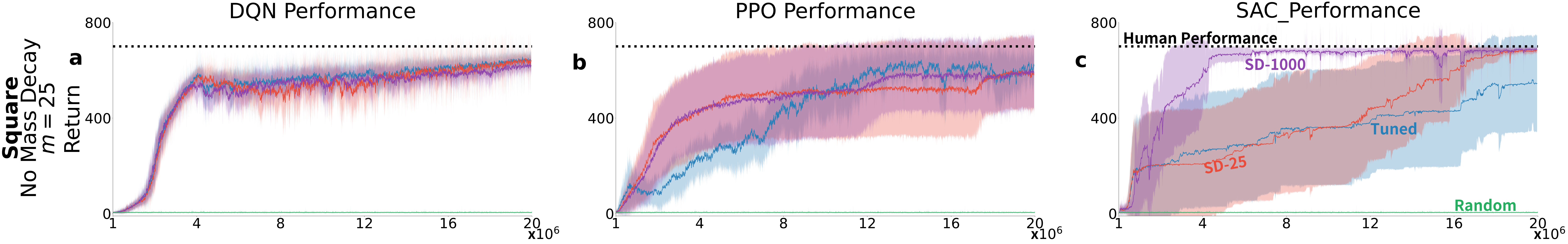}
\includegraphics[width=\textwidth]{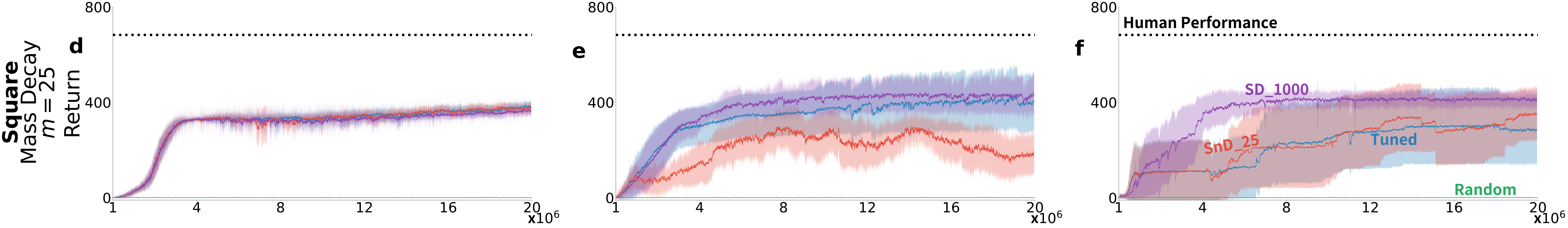}
\includegraphics[width=\textwidth]{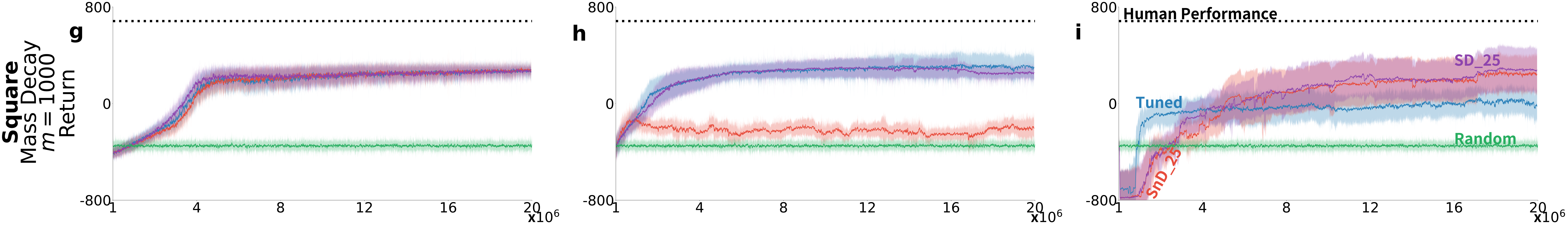}
\includegraphics[width=\textwidth]{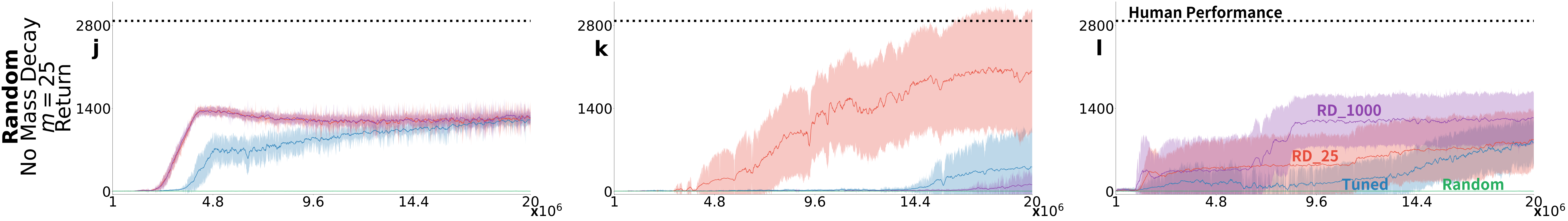}
\includegraphics[width=\textwidth]{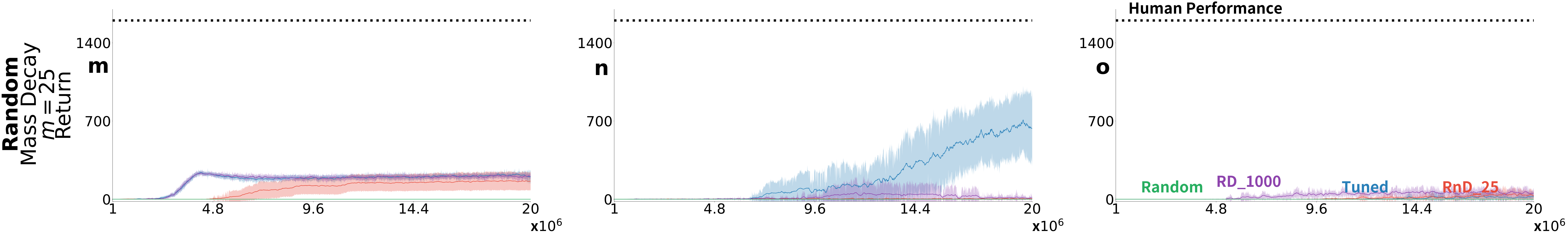}
\includegraphics[width=\textwidth]{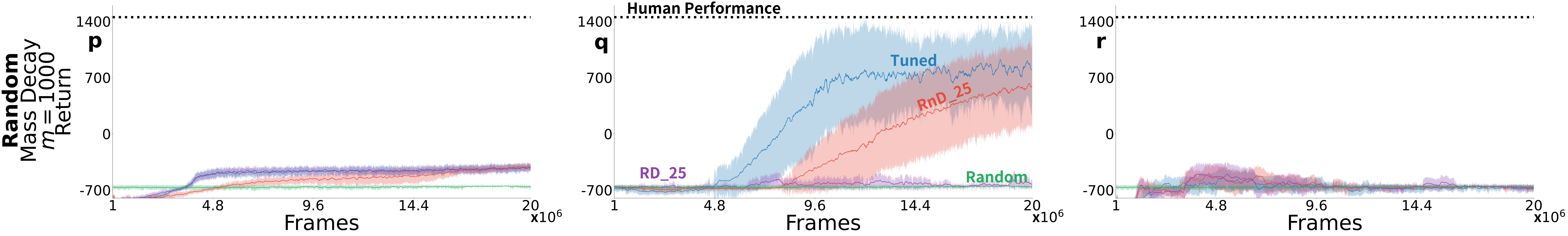}

\caption{Performance of DQN, PPO, and SAC on six mini-games using cross-evaluated hyperparameters. The mini-games are grouped into two categories: the top three rows correspond to the \textbf{square path} group (Square-path pellet collection tasks), and the bottom three rows to the \textbf{random} group (randomly regenerated pellet tasks). For each mini-game, agents are evaluated not only using their own tuned hyperparameters but also using the hyperparameters optimized for the other two tasks within the same group. Each row shows the performance of a baseline algorithm across three hyperparameter configurations. For example, panels \textcircled{a}, \textcircled{b}, and \textcircled{c} show DQN, PPO, and SAC in \textsc{Mini-game} $1$ under different hyperparameter settings. The naming convention is as follows: \texttt{SD-1000} denotes a \texttt{S}quare path setting with mass \texttt{D}ecay and an initial mass of $1000$, while \texttt{SnD-25} indicates \texttt{n}o \texttt{D}ecay and an initial mass of $25$. \texttt{Tuned} is considered the best hyperparameter on a particular mini-game. We use \textit{R} instead of \textit{S} on the panels for the mini-games in which pellets were randomly spread in the environment (instead of a square).}
\label{fig:hypers}
\end{center}
\end{figure}

We ran thousands of GPU jobs to tune and evaluate, on a per-environment basis, the baselines discussed in the previous section. Hyperparameter tuning is a major challenge in RL, and an even bigger one in continual RL. Each algorithm is impacted differently by each hyperparameter. 
 To test robustness, we used hyperparameters tuned in one mini-game to evaluate performance in another. In many tasks, PPO tuned on a particular mini-game often collapses when applied to a different mini-game. For SAC, using hyperparameters tuned for different mini-games sometimes yields better results than the hyperparameters optimized for the task itself! DQN, on the other hand, demonstrates surprising robustness—cross-task hyperparameter transfers have minimal impact on its peak performance. This large variability can be due to the complexities of hyperparameter selection strategies \citep{EmpricalDesignforRL}, including the fact that we could afford only three seeds per configuration, which is certainly not enough for an accurate estimate. This is another common practice in the field that can be quite detrimental. The actual plots for these results are in Figure~\ref{fig:hypers}. These results underscore a critical insight: no single hyperparameter setting is robust across all tasks. 
 \looseness=-1

\subsubsection{PPO augmented with a GRU} \label{sec:experiments_rnns}
One might wonder whether approaches that tackle partial observability would be effective in \texttt{AgarCL}, especially in settings with a high degree of partial observability. As previously discussed, PPO is the best-performing algorithm among those we considered; accordingly, we decided to evaluate PPO augmented with a recurrent network in the pellet-collection \textsc{Mini-Games}.

Specifically, we evaluated PPO with Gated Recurrent Units~\citep[GRUs;][]{cho2014properties}. The model comprises three convolutional layers, followed by a 256-unit GRU, and a final linear layer that splits into actor and critic heads. We re-tuned all hyperparameters for this new algorithm. All other experimental procedures were kept the same.

\begin{figure}[h]
\begin{center}
\includegraphics[width=\textwidth]
{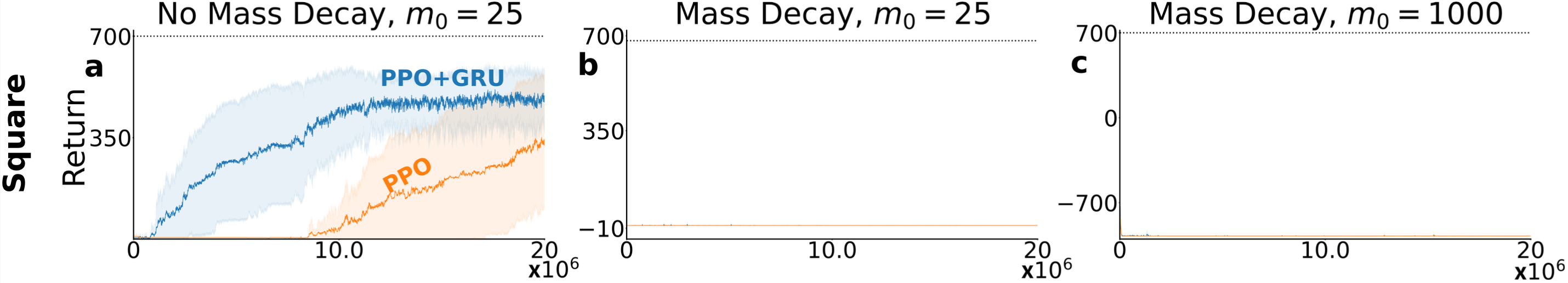}
\includegraphics[width=\textwidth]
{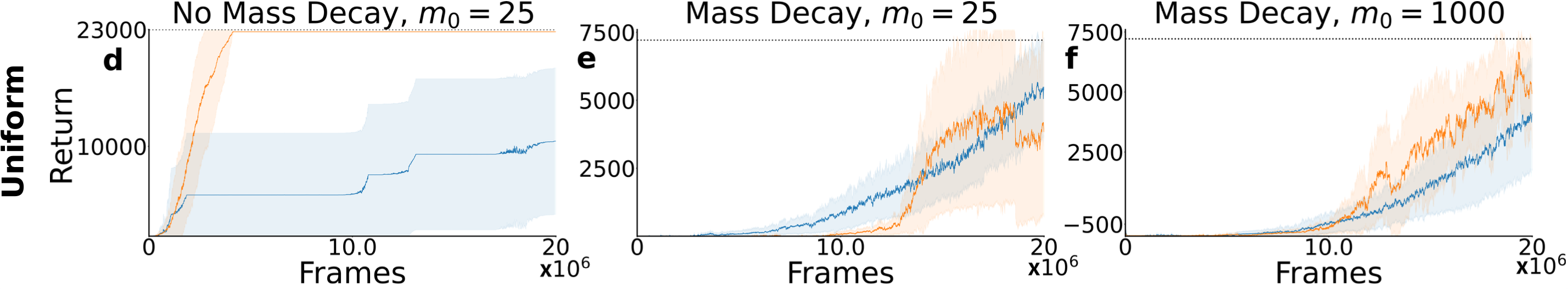}
\caption[Performance of PPO with RNN backbone on \emph{continual} pellet-collection.]{Performance of PPO+GRU and PPO on \emph{continual} pellet-collection mini-games. Panels \textcircled{a}, \textcircled{b}, and \textcircled{c} correspond to the \emph{square-path} tasks (with no mass decay, mass decay $m_0 = 25$, and mass decay $m_0 = 1000$, respectively), while panels \textcircled{d}, \textcircled{e}, and \textcircled{f} correspond to the \emph{uniformly regenerated} pellets with the same decay settings. The shaded regions show the 95\% CI over 10 evaluation runs, computed using the t-distribution. Note that the y-axis scales differ between panels.}
\label{fig:PPO_RNN}
\end{center}
\end{figure}

As shown in Figure~\ref{fig:PPO_RNN}, GRUs did not lead to much improvement. It performed well in relatively simple scenarios, such as \textsc{mini-game} 1 (see panel \textcircled{$a$}), achieving human-level performance. However, it struggled in more challenging tasks, such as mini-games 2 and 3 (panels \textcircled{$b$} and \textcircled{$c$}). Notably, in the uniform variation (panels \textcircled{$d$}-\textcircled{$f$}), the performance of PPO with GRUs was even worse than that of the non-recurrent version. Table~\ref{tab:PPO_RNN_Results} reports the equivalent numerical results.


\begin{table}[H]
  \centering
  \footnotesize
  \resizebox{1.0\textwidth}{!}{%
    \begin{threeparttable}
      \caption[PPO + RNN performance across 6 mini-games]{Performance of PPO with GRU-based recurrent architecture across six mini-games. Results are averaged over the last 100 data points. Values in parentheses denote standard deviations.}
      \label{tab:PPO_RNN_Results}
      \renewcommand{\arraystretch}{1.2}
      \begin{tabular}{@{}l *{6}{c}@{}}
        \toprule
        \textbf{Algorithm} 
          & \textbf{Mini-Game 1} 
          & \textbf{Mini-Game 2} 
          & \textbf{Mini-Game 3} 
          & \textbf{Mini-Game 4} 
          & \textbf{Mini-Game 5} 
          & \textbf{Mini-Game 6} \\
        \midrule
        PPO 
          & 335 \tiny(3.0)  
          & 0 \tiny(0)  
          & $-$975 \tiny(0) 
          & $22{,}463$ \tiny(48.9) 
          & $4{,}018$ \tiny(126.8)  
          & $5{,}064$ \tiny(113.6) \\
          \midrule
        PPO + RNN 
          & 484 \tiny(10.0)  
          & 0 \tiny(0)  
          & $-$974 \tiny(0.3) 
          & $20{,}533$ \tiny(250) 
          & $5{,}329$ \tiny(30.9)  
          & $4{,}036$ \tiny(22.4) \\
          
        \bottomrule
      \end{tabular}
    \end{threeparttable}
  }
\end{table}

\subsection{Interacting with Other Agents}
\label{appendix:other_agents}

Aside from pellet collection, capturing (or fleeing) other agents is a key skill an agent is expected to have in \texttt{AgarCL}. In this section, we evaluate the agent in mini-games where it interacts with a single other bot in the environment. To make the problem easier for the agent, given the difficulty they faced in the continual mini-games, we conducted experiments in an episodic setting, where each episode terminates either when the agent is eaten or after 10,000 time steps, whichever occurs first. We conducted these experiments in two configurations: the \textsc{Large-Dense} and \textsc{Small-Sparse} setups. We discuss both below.

\subsubsection{Large-Dense Setup}
\begin{figure}[h]
\begin{center}
\includegraphics[width=\textwidth]{Figures/hun_Other_Agents_350x350.pdf}
\includegraphics[width=\textwidth]{Figures/Agg_Other_Agents_3_350x350.pdf}
\caption{Performance of DQN, PPO, and SAC on \emph{episodic} other-agent mini-games. Panels \textcircled{a} and \textcircled{d} show the performance of \emph{DQN}, Panels \textcircled{b} and \textcircled{e} show the performance of \emph{PPO}, and Panels \textcircled{c} and \textcircled{f} show the performance of \emph{SAC}. The top row (\textcircled{a}–\textcircled{c}) corresponds to the \emph{hungry-bot} task (mini-game 7), while the bottom row (\textcircled{d}–\textcircled{f}) corresponds to the \emph{aggressive-bot} task (mini-game 8). Each experiment features a fixed-policy bot—either hungry or aggressive—in a standard-sized arena ($350 \times 350$) with 500 pellets. Both the agent and the fixed-policy bot start with an initial mass of 25. Shaded regions represent the 95\% CI over 10 independent runs, computed using the $t$-distribution.}
\label{fig:large_other_agent}
\end{center}
\end{figure}
This first setup uses a standard number of pellets (500) and a normal-sized arena ($350 \times 350$). We used two types of fixed-policy bots: the \textit{hungry} bot, which focuses solely on collecting pellets, and the \textit{aggressive} bot, which prioritizes consuming any entity smaller than itself.

Figure~\ref{fig:large_other_agent} depicts the obtained results as well as the return obtained by the fixed-policy bot, as its return in comparison to the learning agent is quite informative. We can see that both DQN and SAC struggled against the aggressive, hungry bots. Gameplay videos recorded at various evaluation checkpoints revealed that these agents frequently became stuck in corners, a failure mode also observed in pellet collection mini-games. Consequently, the \textit{aggressive} bot could quickly consume the learning agents, as reflected by the early plateaus in their return curves in panels \textcircled{d} and \textcircled{f}. The performance of the \textit{hungry} bot in panels \textcircled{a} and \textcircled{c} appears slightly better, focusing on collecting pellets rather than following the learning agents. The PPO agent exhibited more robust behaviour. Panel \textcircled{e} shows that PPO learned to avoid the aggressive bot effectively while collecting many pellets. This evasive strategy contributed to the steady increase observed in the agent’s return, albeit at a slower rate than its adversary. However, in the hungry setting (panel \textcircled{b}), PPO failed to consume the bot, even though the hungry bot did not attempt to attack it. This suggests that while PPO is effective at survival, it may not exploit opportunities to eliminate non-aggressive opponents.

These results raise the question of whether PPO’s success was due to the large arena and high pellet density. The \textsc{Small-Sparse} setting, discussed next, addresses this question directly.

\subsubsection{Small-Sparse Setup}

\begin{figure}[h]
\begin{center}
\includegraphics[width=\textwidth]{Figures/hun_Other_Agents_200x200.pdf}
\includegraphics[width=\textwidth]{Figures/Agg_Other_Agents_3__200x200.pdf}
\caption[Performance of RL baselines—DQN, PPO, and SAC—on \emph{episodic} other-agent mini-games.]{Performance of RL baselines—DQN, PPO, and SAC—on \emph{episodic} other-agent mini-games. Panels \textcircled{a} and \textcircled{d} show the performance of \emph{DQN}, Panels \textcircled{b} and \textcircled{e} show the performance of \emph{PPO}, and Panels \textcircled{c} and \textcircled{f} show the performance of \emph{SAC}. The top row (\textcircled{a}–\textcircled{c}) corresponds to the \emph{hungry-bot} mini-game, while the bottom row (\textcircled{d}–\textcircled{f}) corresponds to the \emph{aggressive-bot} mini-game. Each experiment features a fixed-policy bot—aggressive or hungry—in a limited arena ($200 \times 200$) with 250 pellets. The agent and the fixed-policy bot start with an initial mass of 25. Shaded regions represent the 95\% confidence intervals over 10 independent runs, computed using the $t$-distribution.}
\label{fig:sparse_other_agent}
\end{center}
\end{figure}
 
To limit the agent’s ability to avoid interaction with bots, we evaluated them in a smaller arena ($200 \times 200$) with fewer pellets (250). This configuration increases the likelihood of direct encounters between the learning agent and the fixed-policy bot.

Consistent with previous findings, the DQN and SAC baselines failed to demonstrate effective learning, irrespective of the type of bot in the environment (see Figure~\ref{fig:sparse_other_agent} on the next page). This is primarily attributable to recurring behavioural failures, such as becoming trapped in corners or failing to evade the bot, which persisted under these more restrictive conditions. Interestingly, the PPO agent also failed to learn in this environment, regardless of whether it was paired with the aggressive or the hungry bot. PPO could not consistently evade the aggressive bot or exploit the passive behaviour of the hungry one. In panel \textcircled{e}, PPO’s return clearly plateaus early, while the aggressive bot’s return steadily increases, indicating its ability to repeatedly eat the agent with ease.
Even more interesting is PPO's performance against the hungry bot. Despite the absence of an active threat, PPO failed to take advantage of the opportunity to survive and collect pellets. The consistent flattening of the agent's learning curves across all settings further reinforces this observation, indicating minimal policy improvement over time.

\subsection{Interacting with Viruses} \label{sec:viruses_minigame}

\begin{figure}[b]
    \centering
    \begin{minipage}{0.45\textwidth}
        \centering
        \includegraphics[width=0.65\linewidth]{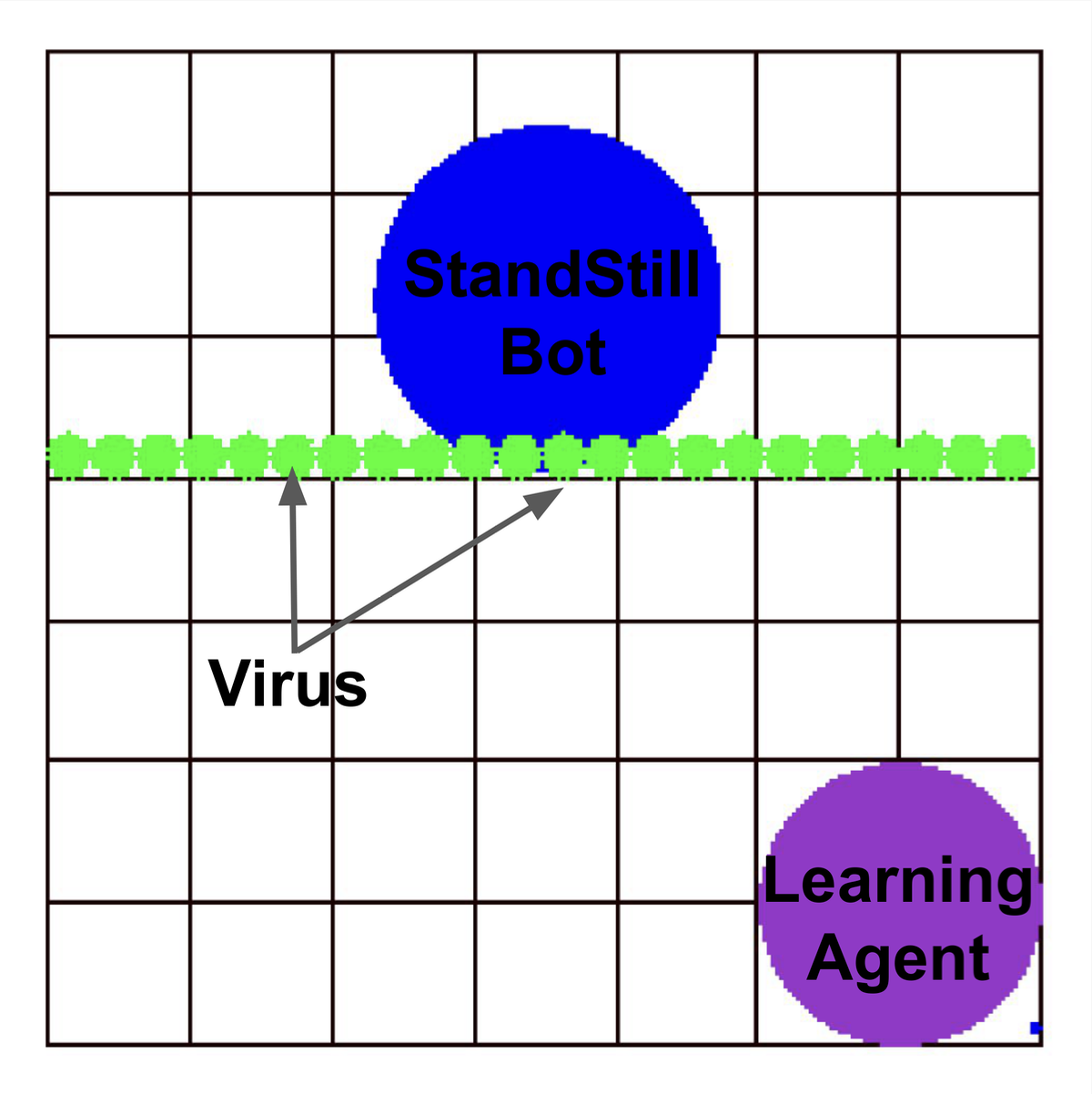}     
        \caption{The learning agent (mass 3${,}$000) is separated from a standstill bot (mass 5${,}$000) by a line of static viruses (mass 100). The arena is fully observable with no pellets, mass decay, or virus respawning.}
        \label{fig:Virus_Mode_11}
    \end{minipage}%
    \hfill
    \begin{minipage}{0.53\textwidth}
        \centering
        \includegraphics[width=\linewidth]{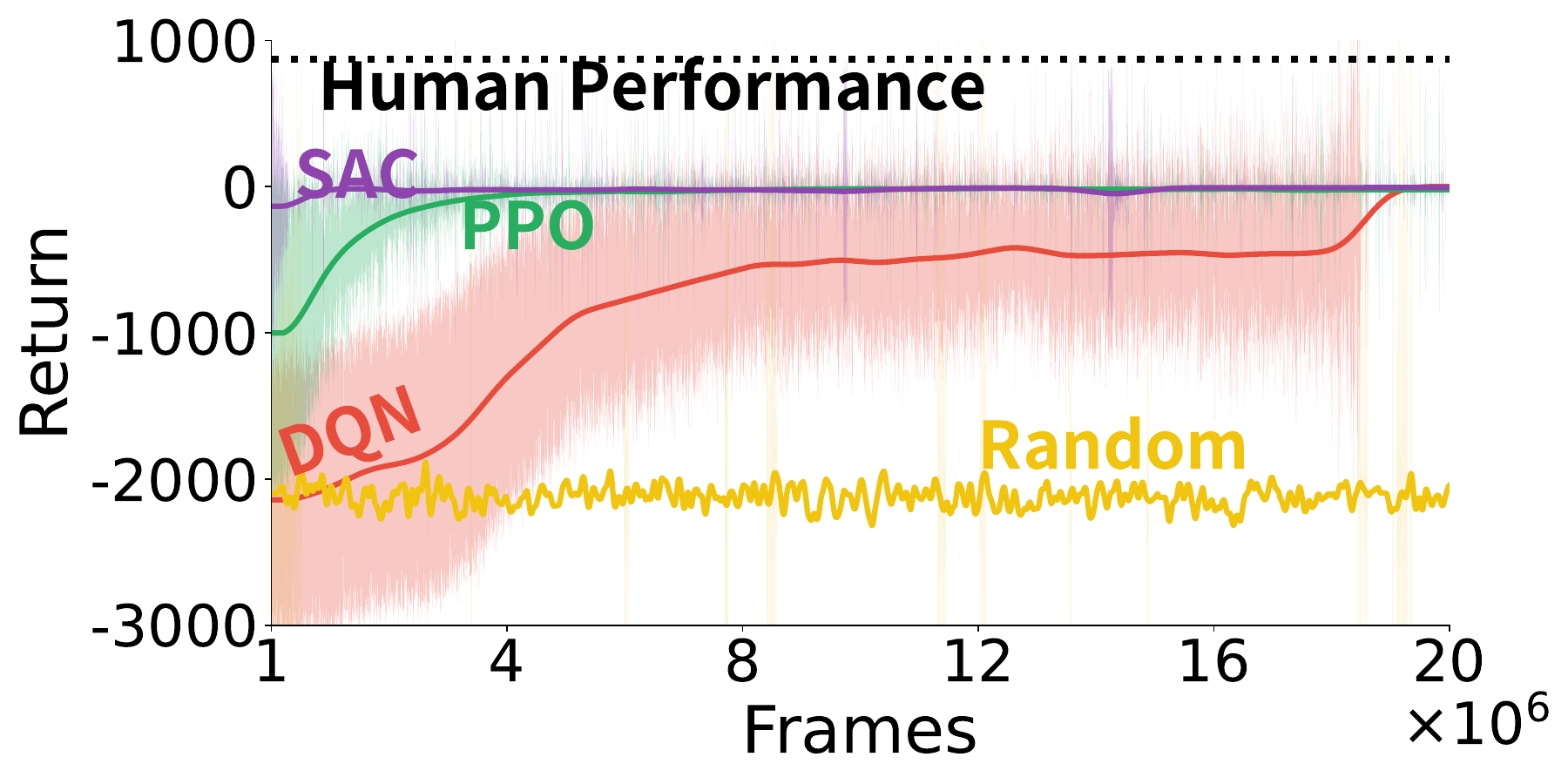}
        \caption{Performance of DQN, PPO, and SAC on the \emph{virus-based} mini-game. Shaded regions show the 95\% CI across 10 runs using the $t$-distribution.}
        \label{fig:Virus_Mode_11_res}
    \end{minipage}
\end{figure}

A final skill we probe for is whether agents can learn how to leverage viruses in the environment, in the most benign setting possible. The mini-game is configured to be fully observable. As illustrated in Figure~\ref{fig:Virus_Mode_11}, the opposing bot remains stationary, and a clear, linear arrangement of viruses is positioned between the agent and the bot. Notably, there are no pellets in this mini-game. The agent begins with a mass of $3000$ and must interact with a larger, stationary bot with a mass of $5000$. To simplify dynamics further, mass decay is disabled. Training in this mini-game follows an episodic setup: an episode ends when one agent is eaten or after $1000$ time steps.

The agent should learn to eject pellets toward the viruses, triggering them to split the stationary bot. The agent can then pass through the virus field and consume the smaller bot fragments. As shown in Figure~\ref{fig:Virus_Mode_11_res}, all learning agents failed to discover this strategy. Instead, they converged on a passive behaviour, remaining stationary throughout the episode. This highlights the difficulty these agents face in learning to leverage environmental elements, such as viruses, for strategic interaction, even in a simplified, fully observable setting.

\section{Numerical Results on Mini-Games and the Full Game}
\label{appendix:Table_Results}

Table~\ref{table_results} summarizes the results of the experiments conducted in this paper. Each reported value represents the final performance, calculated as the average reward over the last 100 evaluation episodes. \looseness=-1

\begin{table}[H]
  \centering
  \footnotesize 
  \resizebox{1.06\textwidth}{!}{%
    \begin{threeparttable}
      \caption{Performance across \texttt{AgarCL} minigames averaged over 10 independent runs (std. dev. is reported between parentheses). Results are averaged over the last 100 data points: episodes in episodic tasks and 100 steps in continual tasks. \looseness=-1} \label{table_results}

      \renewcommand{\arraystretch}{1.2}
      \begin{tabular}{@{}llc *{5}{c}@{}}
        \toprule
        \textbf{Category} 
          & \textbf{Mini-game} 
          & \textbf{Scenarios} 
          & \textbf{DQN} 
          & \textbf{PPO} 
          & \textbf{SAC} 
          & \textbf{Human} 
          & \textbf{Random} \\
        \midrule

        \multirow{6}{*}{\shortstack[l]{Pellet Collection\\(Episodic)}} 
          & 1 & — & 642 \tiny(14.2)  & 605 \tiny(25.3)   & 546 \tiny(6.0)    & 700 \tiny(0)    & 4.81 \tiny(1.3) \\
          & 2 & — & 382 \tiny(9.1)   & 398 \tiny(12.9)   & 285 \tiny(9.8)    & 682 \tiny(0)    & 0.81 \tiny(0.7) \\
          & 3 & — & 271 \tiny(23.1)  & 298 \tiny(28.8)   & $-$16 \tiny(30.3) & 612 \tiny(0)    & $-$356 \tiny(22.7) \\
          & 4 & — & 1189 \tiny(44.2) & 391 \tiny(36.3)   & 790 \tiny(98.6)   & 2876 \tiny(0)   & 4 \tiny(1.3) \\
          & 5 & — & 214 \tiny(13.2)  & 650 \tiny(58.9)   & 16 \tiny(8.4)     & 1600 \tiny(0)   & 0.17 \tiny(0.16) \\
          & 6 & — & $-$426 \tiny(20.0) & 812 \tiny(118.3) & $-$681 \tiny(12.8) & 1452 \tiny(0) & $-$657 \tiny(6.9) \\
        \midrule

        \multirow{9}{*}{\shortstack[l]{Pellet Collection\\(Continual)}} 
          & 1 & — & 619 \tiny(7.9)     & 335 \tiny(3.0)   & 419 \tiny(0.2)     & 700 \tiny(0)    & 0.01 \tiny(0.1) \\
          & 2 & — & 0 \tiny(0)         & 0 \tiny(0)       & 2 \tiny(1.7)       & 682 \tiny(0)    & 0 \tiny(0) \\
          & 3 & — & $-$975 \tiny(0)    & $-$975 \tiny(0)  & $-$975 \tiny(0)    & $-$480 \tiny(0) & $-$975 \tiny(0) \\
          & 4 & Sparse\tnote{1} & —    & 22463 \tiny(48.9)& —                   & —               & — \\
          &   & Dense\tnote{1}  & 21402 \tiny(762.8) & 21970 \tiny(277) & 21997 \tiny(595.6) & 23000 \tiny(0) & 112.2 \tiny(12.5) \\
          & 5 & Sparse & —             & 144 \tiny(10.8)  & —                   & —               & — \\
          &   & Dense  & 132 \tiny(9.3)& 4018 \tiny(126.8)& 0.012 \tiny(0.9)    & 7215 \tiny(0)   & 0.05 \tiny(0.09) \\
          & 6 & Sparse & —             & $-$618.3 \tiny(1.4) & —                & —               & — \\
          &   & Dense  & $-$828 \tiny(10.5) & 5064.4 \tiny(113.6) & $-$968 \tiny(0.01) & 7215 \tiny(0) & $-$973 \tiny(1.1) \\
        \midrule

        \multirow{2}{*}{\shortstack[l]{Other Agent\\(Episodic)}} 
          & $7$---Small, Sparse\tnote{2} & — & 157 \tiny(0.7)   & 253 \tiny(35.6)  & 31 \tiny(0.2)     & 1980 \tiny(0)   & 0.015 \tiny(0.04) \\
          & $7$---Large, Dense\tnote{2}  & — & 146 \tiny(48.4)  & 429.3 \tiny(136.4) & 25 \tiny(14.2) & 2385 \tiny(0)   & 0.1 \tiny(0.16) \\
          & $8$---Small, Sparse & — & 189 \tiny(6.03)   & 267 \tiny(26)  & 114 \tiny(18.8)     & 1800 \tiny(0.015)   & 0.04 \tiny(0.04) \\
          & $8$---Large, Dense  & — & 191 \tiny(14.56)  & 1037 \tiny(55.2) & 69 \tiny(8.9) & 2035 \tiny(0.015)   & 0.04 \tiny(0.04)\\
        \midrule
        
        \multirow{1}{*}{Virus (Episodic)} 
          & $9$ & — & -36 \tiny(4.23) & -46 \tiny(6.45) & 	-5 \tiny(0.1) & 870 \tiny(20.5)   & 	-2319 \tiny(8.94)\\
        \midrule

        Full Game (Grand Arena) & — & — & 4 \tiny(3.1) & 8 \tiny(8.3) & 22 \tiny(11.1) & — & 0.06 \tiny(0.18) \\
        \bottomrule
      \end{tabular}
      \begin{tablenotes}
        \footnotesize
        \item[1] In continuing pellet‐collection tasks, ``Sparse'' refers to an arena with 250 pellets. ``Dense'' uses 500 pellets.
        \item[2] ``Small'' arena: $200\times200$ with 200 pellets. ``Large'' arena: $350\times350$ with 500 pellets.
      \end{tablenotes}
    \end{threeparttable}
  }
\end{table}

\section{How the Number of Pellets Impacts the Agent's Performance}\label{appendix: PPO_Performance_sparse_env}

In this section, we further evaluate how PPO behaves in more challenging pellet-collection mini-games, where the number of available pellets is significantly reduced. 

In Figure~\ref{fig:uniform_pellets_cont}, PPO demonstrates strong performance in collecting pellets, even in the continual setting. However, the task remains relatively easy due to the high number of pellets (500), which allows the agent to find and consume pellets with minimal effort, even in the presence of mass decay. The situation changes when the number of pellets is reduced to 250. Interestingly, PPO failed to learn effectively in \textsc{ mini-games} \textcircled{5} and \textcircled{6}, suggesting increased difficulty with exploration. 

\begin{wrapfigure}[19]{r}{0.5\textwidth}
  \includegraphics[width=0.5\textwidth]{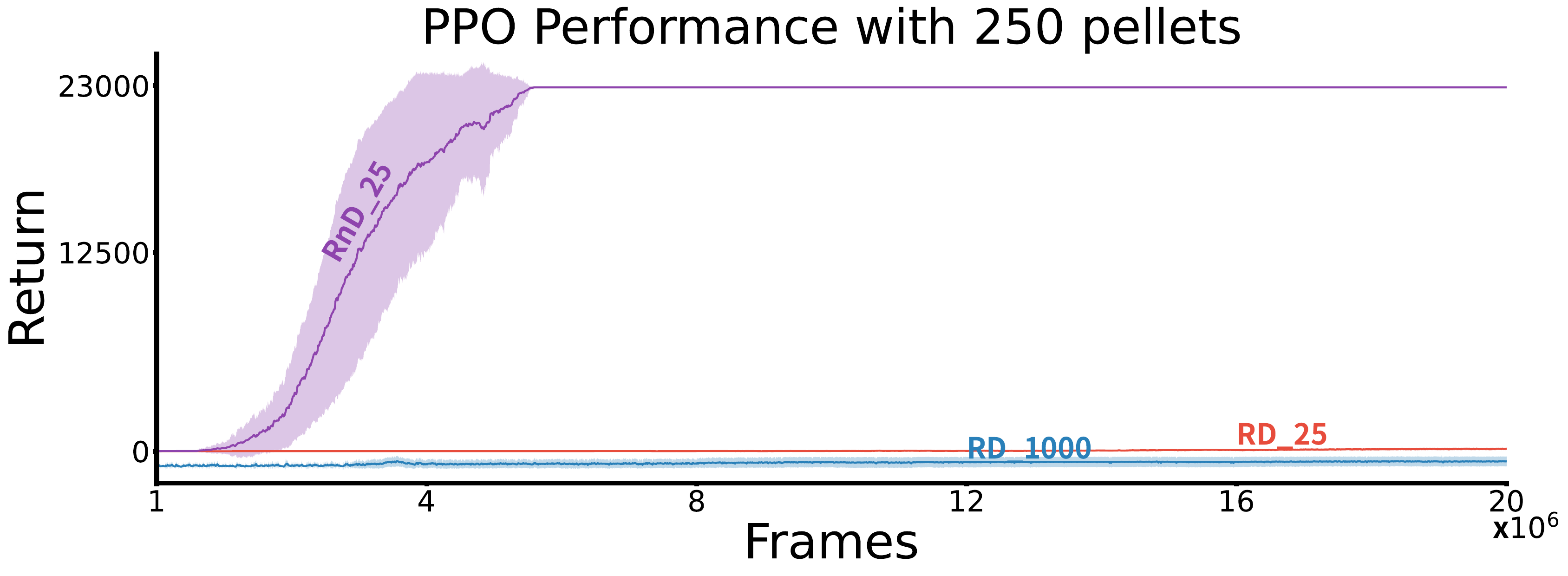}
\caption{PPO performance on \emph{continuing randomly regenerated pellet-collection} in \textsc{mini-games} 4, 5, and 6. \texttt{RD-1000} denotes a \texttt{R}andomly regenerated pellet setting with mass \texttt{D}ecay and an initial mass of $1000$; \texttt{SnD-25} indicates \texttt{n}o \texttt{D}ecay and an initial mass of $25$. The plateau in the \texttt{RnD-25} curve after approx. 1M steps is due to the agent reaching the maximum cumulative reward in \textsc{mini-game}~\textcircled{4}. Shaded regions indicate the 95\% CI over 10 independent runs, using reference values from the t-table.\looseness=-1}

\label{fig:PPO_Performance_with_Limited_Pellets}
\end{wrapfigure}

More generally, following the discussion on Section~\ref{sec:experiments}, in the context of designing easier settings in the full game such that we can see positive learning, we designed a variant of the environment in
which the number of bots was decreased from $8$ to $4$, and we evaluated the impact of having fewer (400) or more (600, 1024) pellets than in the default setup. We maintained 10 viruses in the environment. We evaluated each agent for 86 million training steps. 

\begin{wrapfigure}[16]{r}{0.5\textwidth}
    \centering
\includegraphics[width=0.5\textwidth]{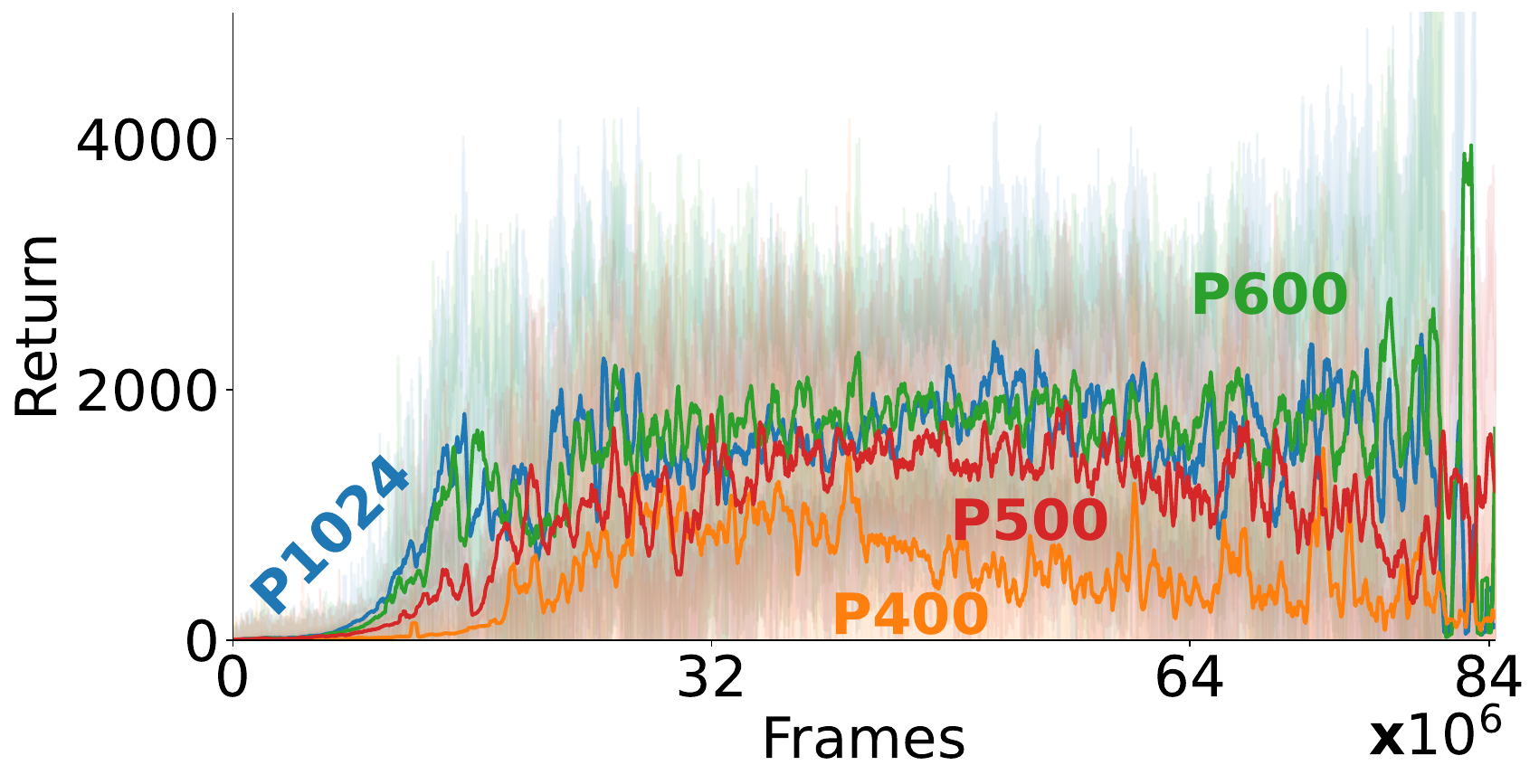}
    \caption[PPO performance across varying pellet densities.]{PPO performance across varying pellet densities within a simplified version of the default setup. All experiments use 10 viruses and 4 bots; with a varying number of pellets, which is labeled in each curve. For example, \texttt{P1024} denotes 1024 pellets. Results are averaged over 10 random seeds, and each curve is smoothed using a moving average with a window of 1000 steps.} \label{fig:PPO_performance_different_density}
\end{wrapfigure}

As illustrated in Figure~\ref{fig:PPO_performance_different_density}, increasing the number of pellets and reducing the number of bots does indeed make the task easier. Some of the curves depict a drop in performance later in training. While this is not surprising in continual RL due to issues such as loss of plasticity~\citep{abbas2023loss,lewandowski2023curvature,dohare2024loss}, we did not investigate this phenomenon further nor evaluate any mitigation strategies. The important observation, despite the high variance across runs, is that this setting makes the problem more tractable, allowing us to consider the impact of freezing the agent's policy. \looseness=-1

These results motivated us to use $1,024$ pellets in the experiment in which we froze the agent's policy (in an easier environment so we could first observe learning).

\section{Continual Learning in \texttt{AgarCL}}\label{app:continual_learning}

In this section, we examine \texttt{AgarCL} from the perspective of continual RL. We first test the impact of freezing the agent's policy in a different setting; then, we evaluate the performance of PPO with continual backpropagation~\citep{dohare2024loss}, an approach explicitly designed for continual learning. \looseness=-1

\subsection{Fixed Policies vs Learning agent in \texttt{AgarCL}}
In our earlier analysis, we identified performance degradation when training was interrupted at $32$ million and $48$ million steps in the setting with $8$ bots and $1,024$ pellets. We also evaluated the impact of freezing the agent's policy in other scenarios. Specifically, we performed an additional experiment under the same conditions but with $4$ bots and $500$ pellets. We can also observe policy collapse for the agents frozen after 32M steps. We conjecture we would also observe policy collapse for the agent frozen after 48M steps if we waited long enough, but if not, this would be evidence that simpler settings (e.g., four bots) might make it easier for the agent to learn some non-trivial policy and maintain it, even though it is quite suboptimal.

\clearpage

\begin{wrapfigure}[15]{r}{0.48\textwidth}
    \centering
    \includegraphics[width=0.46\textwidth]{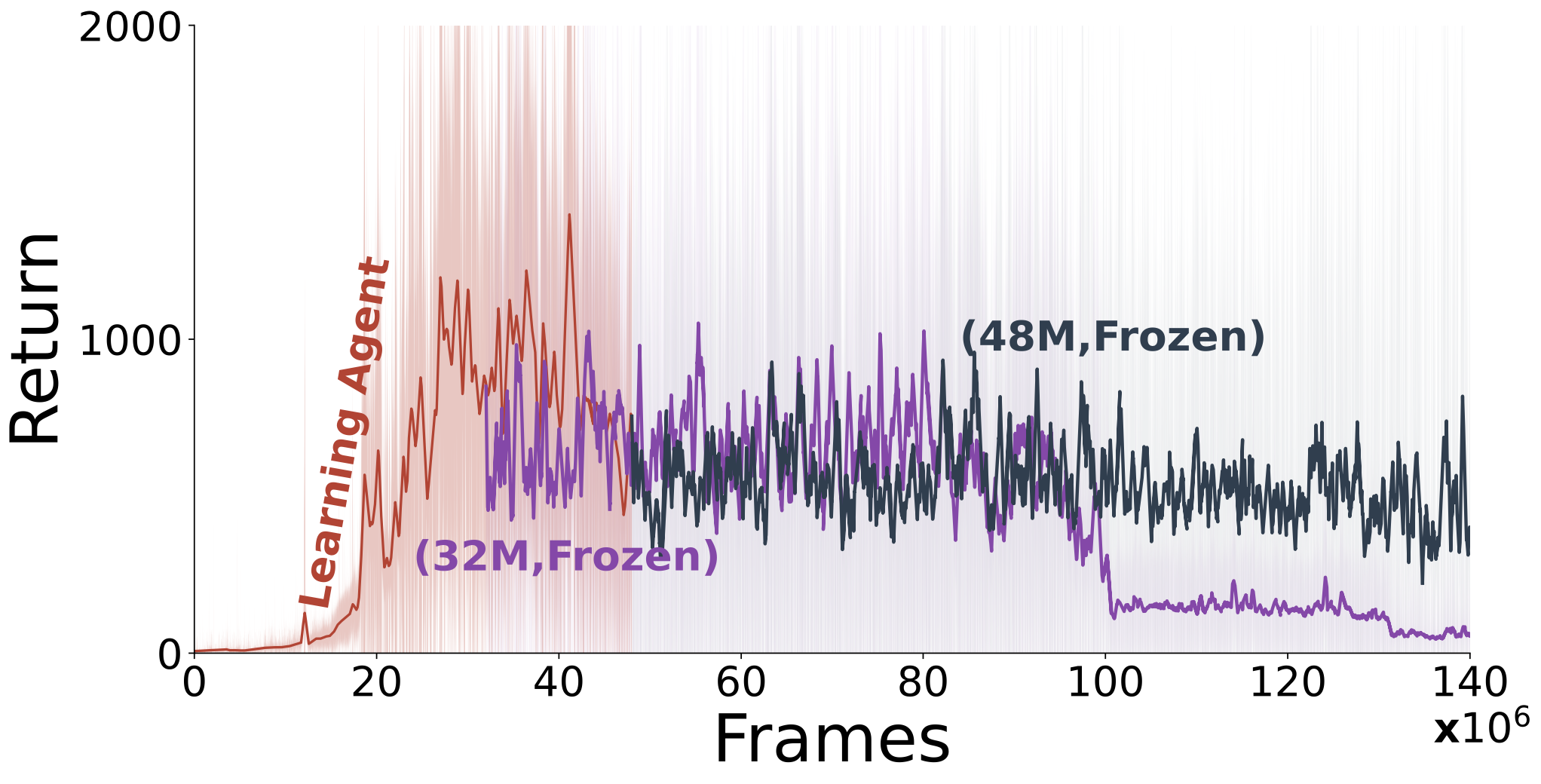}
    \caption{Performance of fixed-policy agents initialized from checkpoints at 32M and 48M steps. We report the moving average over 10 random seeds with a window size of 1000 steps.}
\label{fig:PPO_performance_different_continual_fixed2}
\end{wrapfigure}

\subsection{Details about Experiments with Continual Learning Algorithms}

The algorithms evaluated in Section~\ref{app:continual_backprop} augmented PPO with 
Shrink and Perturb~\citep{ash2020warm}, ReDo~\citep{sokar2023dormantneuronphenomenondeep}, 
or Continual Backpropagation~\citep{dohare2024loss}. In these three approaches, we used the same hyperparameters we used in PPO. We set the algorithm-specific hyperparameters according to the recommendations in the original papers. We detail those below. \looseness=-1
\paragraph{Shrink and Perturb.}
We set $\alpha = 0.4$ (shrink), and $\beta = 0.2$ (perturb), so that, when appropriate, each parameter is updated according to $
x' = 0.4 \, x + 0.2 \, x_0,$
where $x$ is the current weight and $x_0$ is a freshly initialized parameter. 
Shrink-and-Perturb is applied every 500{,}000 time steps.

\paragraph{ReDo.}
ReDo identifies and reinitializes inactive neurons in each linear or convolutional layer (excluding the output layer). For a representative minibatch, we record post-activation values, $a$, and compute per-unit activity scores, $s_j = \mathrm{mean}(|a_j|)$, normalized as $\tilde s_j = s_j / (\bar{s} + 10^{-9})$. Units with $\tilde s_j \le \tau$ are considered dormant. By default, we set $\tau = 0.05$.

ReDo (1) partially reinitializes incoming weights using LeCun normal or Kaiming uniform, (2) zeros outgoing connections to prevent contribution of dormant units, and (3) resets optimizer moments for affected parameters to avoid immediate re-amplification. The operation occurs every 2 million time steps and is computationally cheap compared with full retraining. \looseness=-1

\paragraph{Continual Backpropagation.} Continual Backpropagation (CB) tracks per-feature utility by monitoring activations and output weights, and periodically reinitializes features whose utility falls below a threshold. Replacement is performed via backward-pass hooks: fractional replacement counts accumulate until an integer replacement occurs, and reinitialized input weights are drawn uniformly within layer-specific bounds; corresponding output weights and normalization statistics are reset. Table~\ref{tab:cb_hyperparams} shows our default values.

\begin{table}[h]
\centering
\caption{Continual Backpropagation hyperparameters.}
\begin{tabular}{lcc}
\toprule
Layer type & Replacement rate & Maturity threshold \\
\midrule
Convolutional & $1\times10^{-5}$ & 1000 steps \\
Linear        & $1\times10^{-4}$ & 100 steps \\
\bottomrule
\end{tabular}
\label{tab:cb_hyperparams}
\end{table}

\subsection{Full-Game Results with Continual Learning Methods}

\begin{table}[H]
\centering
\footnotesize
\caption{Full-Game performance over the last 100 steps, averaged across 10 runs. Mean and standard deviation are reported for each algorithm.}
\label{table:full_game_compact}
\begin{tabular}{lcccc}
\toprule
Metric & Continual Backprop & ReDo & Shrink \& Perturb & PPO \\
\midrule
Mean & 3675.8 & 4233.9 & 2074.2 & 762.58 \\
Std  & 4381.81    & 5040.74    & 1868.18  & 1013.76\\
\bottomrule
\end{tabular}
\end{table}

\end{document}